\documentclass[acmtog, nonacm=true]{acmart}
\acmSubmissionID{207}

\usepackage{booktabs} 
\usepackage{rotating}
\usepackage{graphicx}
\usepackage{adjustbox}
\usepackage{arydshln}
\usepackage[ruled]{algorithm2e} 

\SetAlFnt{\small}
\SetAlCapFnt{\small}
\SetAlCapNameFnt{\small}
\SetAlCapHSkip{0pt}

\setcopyright{rightsretained}
\acmJournal{TOG}
\acmYear{2021}\acmVolume{40}\acmNumber{4}\acmArticle{44}\acmMonth{8}
\acmDOI{10.1145/3450626.3459765}

\begin{document}

\title{PhotoApp: Photorealistic Appearance Editing of Head Portraits}

%
\author{Mallikarjun B R}
\email{mbr@mpi-inf.mpg.de}
\author{Ayush Tewari}
\email{atewari@mpi-inf.mpg.de}
\affiliation{%
   \institution{MPI for Informatics, SIC}
   \country{Germany}}
\author{Abdallah Dib}
\email{abdallah.dib@interdigital.com}
\affiliation{%
   \institution{InterDigital R\&I}
   \country{France}}
\author{Tim Weyrich}
\email{t.weyrich@cs.ucl.ac.uk}
\affiliation{%
   \institution{University College London}
   \country{UK}}
\author{Bernd Bickel}
\email{bernd.bickel@ist.ac.at}
\affiliation{%
   \institution{IST Austria}
   \country{Austria}}   
\author{Hans-Peter Seidel}
\email{hpseidel@mpi-sb.mpg.de}
\affiliation{%
   \institution{MPI for Informatics, SIC}
   \country{Germany}}   
\author{Hanspeter Pfister}
\email{pfister@g.harvard.edu}
\affiliation{%
   \institution{Harvard University}
   \country{USA}}  
\author{Wojciech Matusik}
\email{wojciech@csail.mit.edu}
\affiliation{%
   \institution{MIT CSAIL}
   \country{USA}}   
\author{Louis Chevallier}
\email{louis.chevallier@interdigital.com}
\affiliation{%
   \institution{InterDigital R\&I}
   \country{France}}   
 \author{Mohamed Elgharib}
 \email{elgharib@mpi-inf.mpg.de}
 \author{Christian Theobalt}
 \email{theobalt@mpi-inf.mpg.de}
 \affiliation{%
   \institution{MPI for Informatics, SIC}
   \country{Germany}}

\renewcommand\shortauthors{B R, Mallikarjun. et al}

\begin{abstract}
Photorealistic editing of head portraits is a challenging task as humans are very sensitive to inconsistencies in  faces.
We present an approach for high-quality intuitive editing of the camera viewpoint and scene illumination (parameterised with an environment map) in a portrait image. 
This requires our method to capture and control the full reflectance field of the person in the image. 
Most editing approaches rely on supervised learning using training data captured with setups such as light and camera stages.
Such datasets are expensive to acquire, not readily available and do not capture all the rich variations of in-the-wild portrait images.  
In addition, most supervised approaches only focus on relighting, and do not allow camera viewpoint editing. 
Thus, they only capture and control a subset of the reflectance field. 
Recently, portrait editing has been demonstrated by operating in the generative model space of StyleGAN. 
While such approaches do not require direct supervision, there is a significant loss of quality when compared to the supervised approaches. 
In this paper, we present a method which learns from limited supervised training data. 
The training images only include people in a fixed neutral expression with eyes closed, without much hair or background variations. 
Each person is captured under 150 one-light-at-a-time conditions and under 8 camera poses. 
Instead of training directly in the image space, we design a supervised problem which learns transformations in the latent space of StyleGAN. 
This combines the best of supervised learning and generative adversarial modeling.
We show that the StyleGAN prior allows for generalisation to different expressions, hairstyles and backgrounds. 
This produces high-quality photorealistic results for in-the-wild images and significantly outperforms existing methods. 
Our approach can edit the illumination and pose simultaneously, and runs at interactive rates.

\end{abstract}

%

%
%

%

\begin{teaserfigure}
   \centering
   \includegraphics[width=\textwidth]{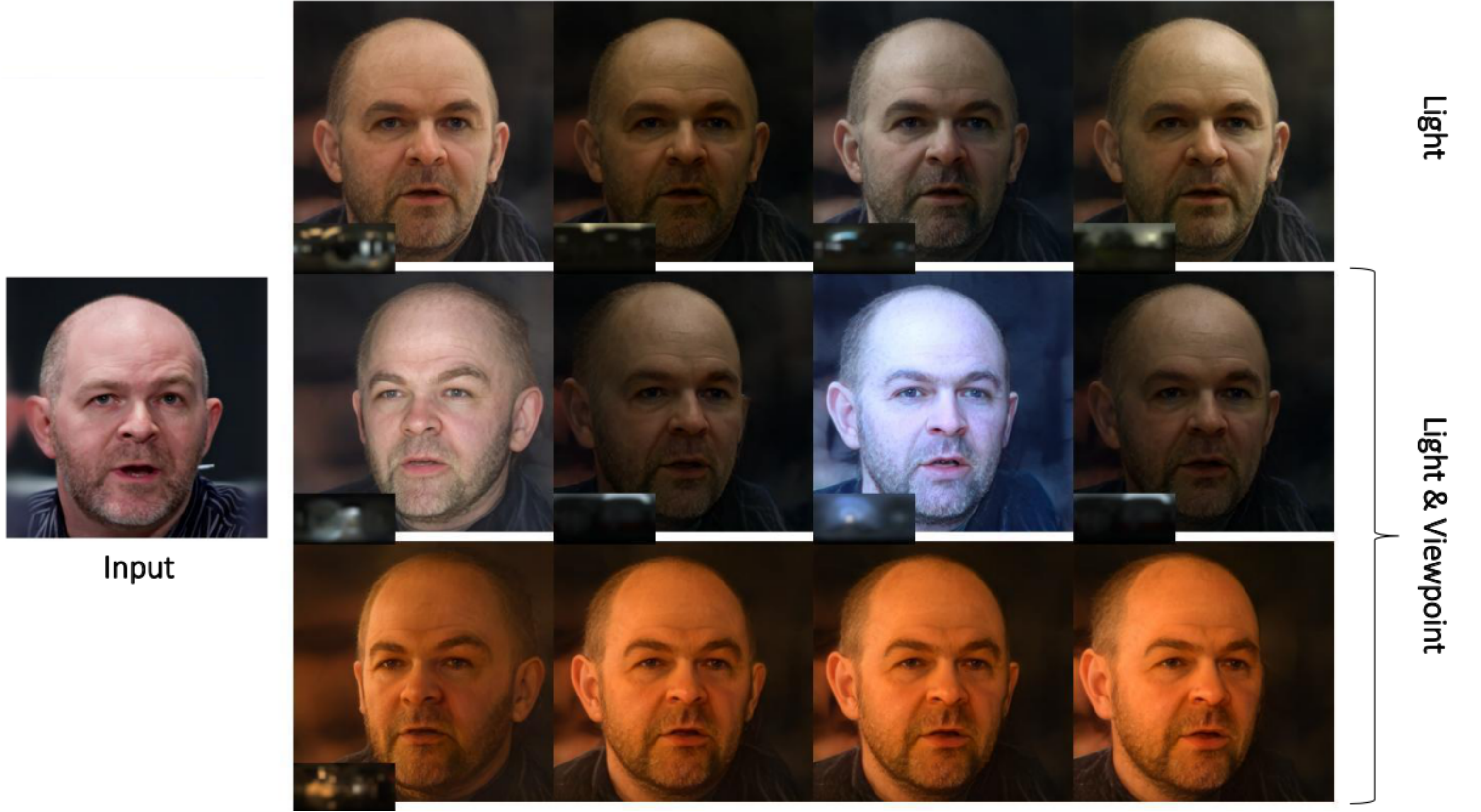}
   \caption{We present a method for high-quality appearance editing of head portraits. Given an input image, our approach edits its appearance using a target environment map (see insets), and a target camera viewpoint. 
   We achieve high-quality photorealistic results for in the wild images, capturing a wide variety of reflectance properties.
   Our method is trained on a light-stage dataset, using a combination of supervised learning and generative adversarial modeling which allows for accurate editing as well as generalisation outside the dataset. Portrait images are from \citet{Shih14} and environment maps are from \cite{hold2019deep,gardner2017learning}.}
   \label{fig:teaser}
\end{teaserfigure} 

\maketitle

\section{Introduction}
Portrait photos are among the most important photographic depictions of humans and their loved ones.
Even though the quality of cameras and thus the photographs have improved dramatically, there arises many cases where people would like to change the scene illumination and camera pose after the image has been captured.
Editing the appearance of the image after capture has applications in post-production, casual photography and virtual reality.
Given a monocular portrait image and a target illumination and camera pose, we present a method for relighting the portrait and editing the camera pose in a photorealistic manner. 
This is a challenging task, as the appearance of the person in the image includes complex effects such as subsurface scattering and self-shadowing. 
Changing the camera requires reasoning about occluded surfaces. 
Humans are very sensitive to inconsistencies in portrait images, and a high level of photorealism is necessary for convincing editing. 
This requires our method to correctly reason about the interactions of the lights in the scene with the surface, and edit them at photorealistic quality. 
We are interested in editing in-the-wild images with a very wide range of illumination and pose conditions. 
We only rely on a single image of an identity unseen during training. 
These constraints make the problem very challenging. 

Several methods have been proposed for editing portrait appearance in the literature. 
One category of methods~\cite{debevec2000acquiring, Weyrich2006Analysis, mvfc_ghosh} address this problem by explicitly modelling the reflectance of the human face~\cite{Kajiya86}.
While these approaches provide well-defined, semantically meaningful reflectance output, they require the person to be captured under multi-view and multi-lit configurations.
They also do not edit the full portrait image, just the inner face region, missing out important portrait components such as hair and eyes. 
Recently, several deep learning-based methods have been proposed for appearance editing. 
These methods use large light-stage datasets which consist of a limited number of people illuminated by different light sources and captured from different camera viewpoints.
A neural network is trained on such datasets which enables inference from a single image. 
Some methods~\cite{yamaguchi_high-fidelity_2018,lattas2020avatarme} regress the reflectance of the face from a monocular image in the form of diffuse and specular components.
Neural representations for face reflectance fields have also been explored recently~\cite{mbr_frf}.
While these methods can work with a single image, they still only model the inner face region, missing out on important details such as hair and eyes.
In contrast to the previous methods, several approaches only capture and edit a subset of the reflectance field. 
These approaches only allow for the editing of either scene illumination or camera pose. 
Most relighting methods directly learn a mapping from the input image to its relit version using a light-stage training dataset~\cite{sipr, nestmeyer2020faceRelighting, sun2020light}.
The controlled setting and limited variety of such datasets limits performance while generalising to in-the-wild images. 
\citet{Zhou_2019_ICCV} attempted to break out from the complexity of capturing light-stage datasets and from their limited variations. 
Instead, they proposed to use a synthetic dataset of in-the-wild images, synthesised with different illuminations. 
Illumination is modeled using spherical harmonics.
The use of synthetic data impacts the photorealism of the results.  
All of these approaches do not allow for changing the camera pose. 
Several methods exist for only editing the camera pose and expressions~\cite{kim2018deep,nagano2018pagan,Siarohin19NeurIPS,Geng18,Wiles18X2Face,Elor17}. 
These methods are commonly trained on videos. 
While person-specific methods~\cite{kim2018deep,thies2019deferred} can obtain high-quality results, methods which generalise to unseen identities~\cite{Siarohin19NeurIPS,Wiles18X2Face} are limited in terms of photorealism. 
In addition, none of them can edit the scene illumination.

Recently, \citet{tewari2020pie} proposed Portrait Image Embedding (PIE), an approach for editing the illumination and camera pose in portrait images by leveraging the StyleGAN generative model~\cite{stylegan}.
PIE computes a StyleGAN embedding for the input image which allows for editing of various face semantics. 
As StyleGAN represents a manifold of photorealistic portraits, PIE can edit the full image with high quality.
However, due to the absence of labelled data, the supervision for the method is defined using a 3D reconstruction of the face. 
This supervision is indirect and not over the complete image, leading to  results that still lack sufficient accuracy and photorealism. 
It uses a low-dimensional representation of the scene illumination and can thus not synthesize results with higher-frequency lights. 
Furthermore, PIE solves a computationally expensive optimisation problem  taking several minutes to compute the embedding.
%

We therefore propose a technique for high-quality intuitive editing of scene illumination and camera pose in a head portrait image. 
Our method combines the best of generative modeling and supervised learning approaches, and creates results of much higher quality compared to previous methods.
We learn to transform the StyleGAN latent code of the input image into the latent code of the output.
We perform this learning in a supervised manner by leveraging a light-stage dataset, containing multiple identities shot from different viewpoints and under several illumination conditions. 
Learning in the StyleGAN space allows us to synthesise photorealistic results for general person identities seen under in-the-wild conditions.
Our method can handle properties such as shadows and other complex appearance, and can synthesise full portrait images including hair, upper body and background. 
We inherit the high photorealism and diversity of the StyleGAN portrait manifold in our solution, which allows us to outperform methods that only use light-stage training data~\cite{sipr}.
Our method has analogies to self-supervised discriminative methods~\cite{jing2020self}. 
We show that the StyleGAN latent representation allows for generalisation even with very little training data. 
We obtain high-quality results of our method even when trained on just 15 identities. 
Our formulation does not make any prior assumptions on the underlying surface reflectance or scene illumination (other than it being distant) and rather directly predicts the appearance as a function of the target environment map and camera pose. 
This leads to significantly more photorealistic results compared to  methods that use spherical-harmonic illumination representations~\cite{Zhou_2019_ICCV,tewari2020pie,abdal2020styleflow} which are limited to only modeling low-frequency illumination conditions.
Furthermore, directly supervising our method using a multi-view and multi-lit light-stage dataset allows us to produce significantly more photorealistic results than PIE~\cite{tewari2020pie}.
Our method can additionally edit at a faster speed, using just a single feedforward pass, and also edit both illumination and pose simultaneously, unlike PIE.
Compared to traditional relighting approaches~\cite{sun2020light,Zhou_2019_ICCV}, we obtain higher-quality results as well as allow for changing the camera pose. 
In summary we make the following contributions: 

\begin{itemize}
 
\item We combine the strength of supervised learning and generative adversarial modeling in a new way to develop a  technique for high-quality editing of scene illumination and camera pose in portrait images. Both properties can be edited simultaneously.
\item Our novel formulation allows for generalisation to in-the-wild images with significantly higher quality results than related methods. It also allows for training with limited amount of supervision. 

\end{itemize}
\section{Related Work}
\begin{figure*}[ht]
   \includegraphics[width=\linewidth]{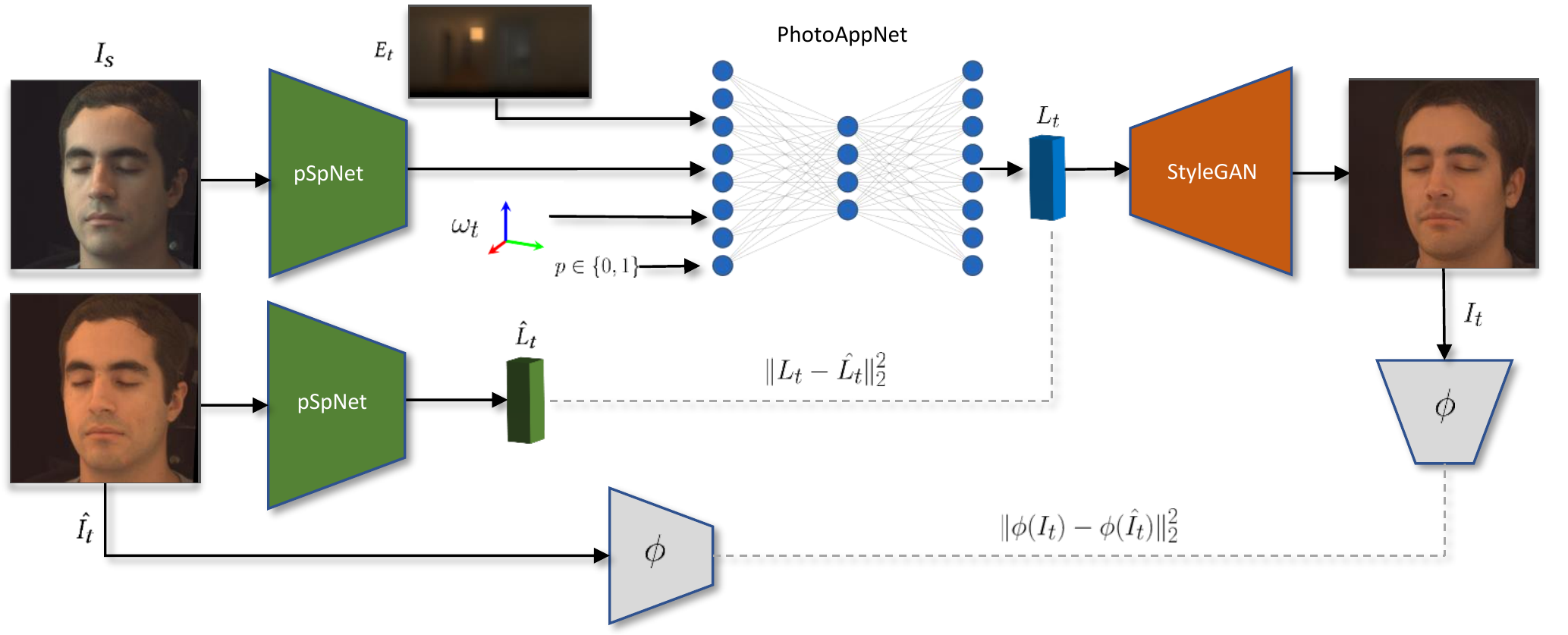}
   \caption{Our method allows for editing the scene illumination $E_t$ and camera pose $\omega_t$ in an input source image $I_s$. 
   We learn to map the StyleGAN~\cite{stylegan2} latent code $L_s$ of the source image, estimated using pSpNet~\cite{richardson2020encoding} to the latent code ${L_t}$ of the output image. 
   StyleGAN~\cite{stylegan2} is then used to synthesis the final output ${I_t}$. 
   Our method is trained in a supervised manner using a light-stage dataset with multiple cameras and light sources.
   For training, we use a latent loss and a perceptual loss defined using a pretrained network $\phi$. 
   Supervised learning in the latent space of StyleGAN allows for high-quality editing which can generalise to in-the-wild images. Portrait images are from \citet{Weyrich2006Analysis} and environment map is from \cite{hold2019deep,gardner2017learning}. 
   }
   \label{fig:pipeline}
\end{figure*}
%

%
In this section we look at related works that can edit the scene parameters in a head portrait image. 
We refer the reader to the state-of-the-art report
of~\citet{tewari2020state} for more details on neural rendering approaches. 

The seminal work of \citet{debevec2000acquiring} introduced a
light-stage apparatus to capture the \emph{reflectance field} 
of a human face, that is, its appearance under a multitude of different
lighting directions. Through weighted superposition of images of the
illumination conditions, their method recreates high-quality images of
the face under any target illumination. By employing additional
cameras and geometry reconstruction, and gathering data from the
additional viewpoints, they further fit a simple bi-directional
radiance distribution function (BRDF) allowing for novel-light
\emph{and} -view renderings of the face. Their method, however, is
limited to reproducing the specific face that was captured.
\citet{Weyrich2006Analysis} extend this concept using a setup with a
much larger number of cameras (16) and a reconstruction pipeline that
extracts geometry and a detailed spatially-varying BRDF (SVBRDF) of a
face. By scanning hundreds of subjects that way, they extract
generalisable statistical information on appearance traits depending
on age, gender and ethnicity. The generative power of the extracted
quantities, however, is heavily constrained, and examples of sematic appearance
editing were limited to subjects from within their face database. In
our work, we revisit their original dataset using a state-of-the-art
learning framework.

Another category of methods tries to infer geometry and reflectance
properties from single, unconstrained images. \citet{Shu17NeuralEditing} 
and~\citet{sfsnetSengupta18} decompose the image into simple intrinsic
components, that is, normals, diffuse albedo and shading. With the
assumption of Lambertian surface reflectance, these methods use
spherical harmonics to model the scene illumination; however, the
starkly simplified assumption ignores perceptually important reflectance
properties which leads to limited photrealism.
Others infer more general surface reflectance, with fewer 
assumptions about incident illumination~\cite{yamaguchi_high-fidelity_2018,
  lattas2020avatarme,mbr_frf}. While such techniques can capture rich
reflectance properties, they do not synthesise the full portrait,
missing out on important components such as hair, eyes and mouth.

Recently, several methods addressed the simplified problem of only relighting a head portrait in the fixed input pose~\cite{sipr_ex, nestmeyer2020faceRelighting, sipr, Zhou_2019_ICCV,zhang2020portrait,Meka19}.
\citet{nestmeyer2020faceRelighting} used a light-stage dataset to train a model that explicitly regresses a diffuse reflectance, as well as a residual component which accounts for specularity and other effects.
Similarly, \citet{sipr_ex} used a light-stage dataset to compute the ground truth diffuse albedo, normal, specularity and shadow images. A network is trained to regress each of these components which are then used in another network to finally relight the portrait image.
Instead of explicitly estimating the different reflectance components, methods such as~\citet{sipr, Zhou_2019_ICCV} directly regress the relighted version of the portrait given the imput image and target illumination.
Here, the target illumination is parameterised either in the form of environment map~\cite{sipr} or spherical harmonics~\cite{Zhou_2019_ICCV}. 
While \citet{sipr} used light-stage data to obtain their ground truth for supervised learning, \citet{Zhou_2019_ICCV} used a ratio image-based approach to generate synthetic training data.

Recently, \citet{zhang2020portrait} proposed a method to remove harsh shadows from a monocular portrait image. They created a synthetic data from in-the-wild images by augmenting shadows and training a network to remove these  shadows. 
Using a light-stage dataset, another network is trained to smooth the artifacts that could remain from the first network.
While the methods of~\cite{sipr_ex, nestmeyer2020faceRelighting, sipr, Zhou_2019_ICCV,zhang2020portrait,Meka19} can produce high-quality relighting results, they either focus on shadow removal~\cite{zhang2020portrait}, or limited by spherical-harmonics illumination representation~\cite{Zhou_2019_ICCV}. 
In addition, methods trained on light-stage or synthetic datasets struggle to generalise to in-the-wild.
They are also limited to only relighting, as they cannot change the camera viewpoint. 

There are several methods for editing the head pose of portrait images~\cite{kim2018deep,nagano2018pagan,Siarohin19NeurIPS,Geng18,Wiles18X2Face,Elor17}. 
While \citet{kim2018deep} require a training video of the examined subject, the techniques of~\citet{nagano2018pagan,Siarohin19NeurIPS,Geng18,Wiles18X2Face,Elor17} can directly operate on a single image. 
However, \citet{nagano2018pagan} does not synthesise the hair and the approaches of~\citet{Siarohin19NeurIPS,Wiles18X2Face} lack explicit 3D modeling and only allow for control using a driving video. 
The approaches of \citet{Elor17,Geng18} rely on warping of the image guided by face mesh deformations, and are thus limited to very small edits in pose. 
%
Furthermore, these approaches can not change the scene illumination. 

Recently, ~\citet{tewari2020pie} proposed PIE, a method which can relight, change expressions and synthesise novel views of the portrait image using a generative model. 
PIE is based on StyleRig~\cite{tewari2020stylerig} which maps the control space of a 3D morphable face model to the latent space of StyleGAN~\cite{stylegan} in a self-supervised manner. 
It further imposes an identity perseverance loss to ensure the source identity is maintained during editing. 
Even though PIE inherits the high photorealism of the StyleGAN portrait manifold, its lack of direct supervision for appearance editing limits its performance and impacts the overall photorealism. 
The scene illumination is parameterised using spherical harmonics as it relies on a monocular 3D reconstruction approach to define its control space. 
Thus, it only allows for rendering using low-frequency scene illumination.
In addition, PIE can not edit the illumination and pose simultaneously, but rather one at a time.
PIE solves an expensive optimisation for the image which is time consuming,  taking around 10 minutes per image on an NVIDIA V100 GPU.
Concurrent to us, \citet{abdal2020styleflow} also propose a method for semantic editing of portrait images using latent space transformations of StyleGAN. 
They use an invertible network based on continuous normalising flows to map semantic input parameters such as head pose and scene illumination into the StyleGAN latent vectors. 
The input parametrisation for the illumination is spherical harmonics like PIE, which limits its relighting capabilities. 
This method is also trained without explicit supervision, i.e., images of the same person with different scene parameters. 
This limits the quality of the results. 
While there are several other approaches which demonstrate transformations of StyleGAN latent vectors for semantic manipulation~\cite{tewari2020stylerig,shen2020interpreting,harkonen2020ganspace,collins2020editing}, these methods focus on StyleGAN generated images, and do not produce high-quality and high-resolution results for real existing images. 

\begin{figure}
    \centering
    \includegraphics[width=0.45\textwidth]{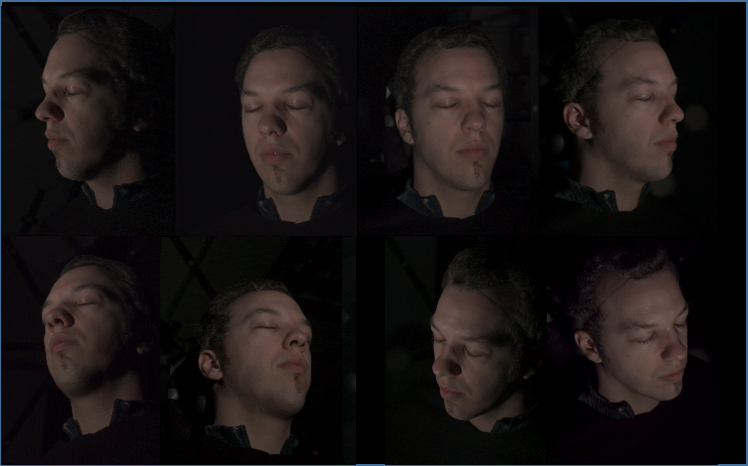}
    \caption{Visualisation of the camera poses in the training dataset. Images are from \citet{Weyrich2006Analysis}.}
    \label{fig:dataset}
\end{figure}
\section{Method}

Our method takes as input an in-the-wild portrait image, a target illumination and the target camera pose. 
The output is a portrait image of the same identity, synthesised with the target camera and lit by the target illumination. 
Given a light-stage dataset of multiple independent illumination sources and viewpoints, the naive approach could be to learn the transformations directly in image space. 
Instead, we propose to learn the mapping in the latent space of StyleGAN~\cite{stylegan2}. %
We show that learning using this latent representation helps in generalisation to in-the-wild images with high photorealism. 
We use StyleGAN2 in our implementation, referred to as StyleGAN for better comprehension. 

\subsection{Dataset}
\label{sec:dataset}
We make use of a light-stage~\cite{Weyrich2006Analysis} dataset for training our solution.
This dataset contains 341 identities captured with 8 different cameras placed in the frontal hemisphere of the face. 
The camera poses available are shown in Fig.~\ref{fig:dataset}.
The dataset also contains 150 light source evenly distributed on the sphere. 
Using this setup, each image is captured with one-light-at-a-time (OLAT) light. 
Given $150$ OLAT images of a person with  a specific camera pose, we can linearly combine them using an environment map to obtain relight portrait images~\cite{debevec2000acquiring}.
We use 205 HDR environment maps from the Laval Outdoor~\cite{hold2019deep} and 2233 from the Naval Indoor~\cite{gardner2017learning} dataset for generating naturally lit images.  
Camera poses for the images are estimated using the approach of \citet{yang2019fsa}. 
Out of the 341 identities, we use 300 for training and the rest for testing.  
We synthesise 300 transformed images for each identity with randomly selected environment maps and camera viewpoints.
Our training set consists of input-ground truth pairs of the same identity along with the target pose and environment map. 
The camera viewpoint of the ground truth is kept identical to the input for quarter of the training data. 
In the remaining, this camera viewpoint is randomly selected. 
The test set includes pairs from the test identities for quantitative evaluations, as well as in-the-wild images for qualitative evaluations, see Sec.~\ref{Sec:results}.

\subsection{Network Architecture}

Fig.~\ref{fig:pipeline} shows an overview of our method. 
Our approach takes as input a source image $I_s$, target illumination $E_t$ and camera pose $\omega_t$, and a binary input $p$.
The value of $p$ is set to $0$ when the target pose is same as that of the input, and $1$ when they are different.
This conditioning input helps in better preservation of the input camera pose for relighting results. 
The ground truth image for training is represented as $\hat{I_t}$.
Camera pose is parameterised using Euler angles. 
We represent the illumination $E_t$ as a $450$ dimensional vectorised environment map.
This corresponds to the 150 RGB discrete light sources.
A core component of our approach is the PhotoAppNet neural network, which maps the input image to the edited output image in the latent space of StyleGAN (see Fig.~\ref{fig:pipeline}).
We first compute the latent representations of $I_s$ and $\hat{I_t}$ as $L_s$ and $\hat{L_t}$ using the pretrained network of \citet{richardson2020encoding} (pSpNet in Fig.~\ref{fig:pipeline}).
The latent space used is $18\times512$ dimensional, corresponding to the W+ space of StyleGAN.
The output of PhotoAppNet is a displacement to the input in the StyleGAN latent space. 
This is then added to the input latent code to compute $L_t$, which is used by StyleGAN to generate the output image $I_t$.
We only train PhotoAppNet, while pSpNet and StyleGAN are pretrained and fixed.

We use an MLP-based architecture with a single hidden layer of length $512$. 
ReLU activation is used after the hidden layer. 
We use independent networks for each of the $18$ latent vectors of length $512$ corresponding to different resolutions.
This is motivated by the design of the StyleGAN network where each $512$ dimensional latent code controls a different frequency of image features. 
The output of each independent network is the output latent code corresponding to the same resolution. 

\subsection{Loss Function}
We use multiple loss terms to train our network.  
\begin{align}
    \mathcal{L}(I_t, L_t, \hat{I_t}, \hat{L_t}, \mathbf{\theta_n}) = &   \mathcal{L}_\text{l}(L_t, \hat{L_t}, \mathbf{\theta_n}) + 
      \mathcal{L}_\text{p}(I_t, \hat{I_t}, \mathbf{\theta_n})
    \enspace{.}
    \label{eq:loss}
\end{align}
Here, {$\theta_n$} denotes the network parameters of PhotoAppNet.
Both terms are weighed equally.
The first term is a StyleGAN latent loss defined as
\begin{align}
\nonumber
    \mathcal{L}_\text{l}(L_t, \hat{L_t}, \mathbf{\theta_n}) =
    \| L_t - \hat{L_t} \|_2^2
    \enspace{.}
\end{align}
It enforces the StyleGAN latent code of the output image ${L_t}$ to be close to the ground truth latent code $\hat{L_t}$. 
%
The second term is a perceptual loss defined as
\begin{align}
\nonumber
    \mathcal{L}_\text{p}(I_t, \hat{I_t}, \mathbf{\theta_n}) = 
    \| \phi(I_t) - \phi(\hat{I_t}) \|_2^2
    \enspace{.}
\end{align}
Here, we employ the learned perceptual similarity metric LPIPS~\cite{zhang2018perceptual}.
An $\ell_2$ loss is used to compare the AlexNet~\cite{krizhevsky2012imagenet} features $\phi()$ of the synthesised output and the ground truth images. 

\subsection{Network Training}
We implement our method in PyTorch and optimise for the weights of PhotoAppNet by minimising the loss function in Eq.~\ref{eq:loss}. 
We use Adam solver with a learning rate of 0.0001 and default hyperparameters.
As mentioned earlier, the  StyleGAN encoder (pSpNet) and generator~\cite{richardson2020encoding,stylegan2} are pretrained  
and fixed during training.
We optimise over our training set samples using a batch size equal to $1$. 
Since in-the-wild images are very different from the light-stage data, it is difficult to assess convergence using a light-stage validation dataset. 
As such, we train our networks using an in-the-wild validation set using qualitative evaluations. 
Our network take around 10 hours to train on a single NVIDIA Quadro RTX 8000 GPU. 
\begin{figure*}
\centering
   \includegraphics[width=0.82\textwidth]{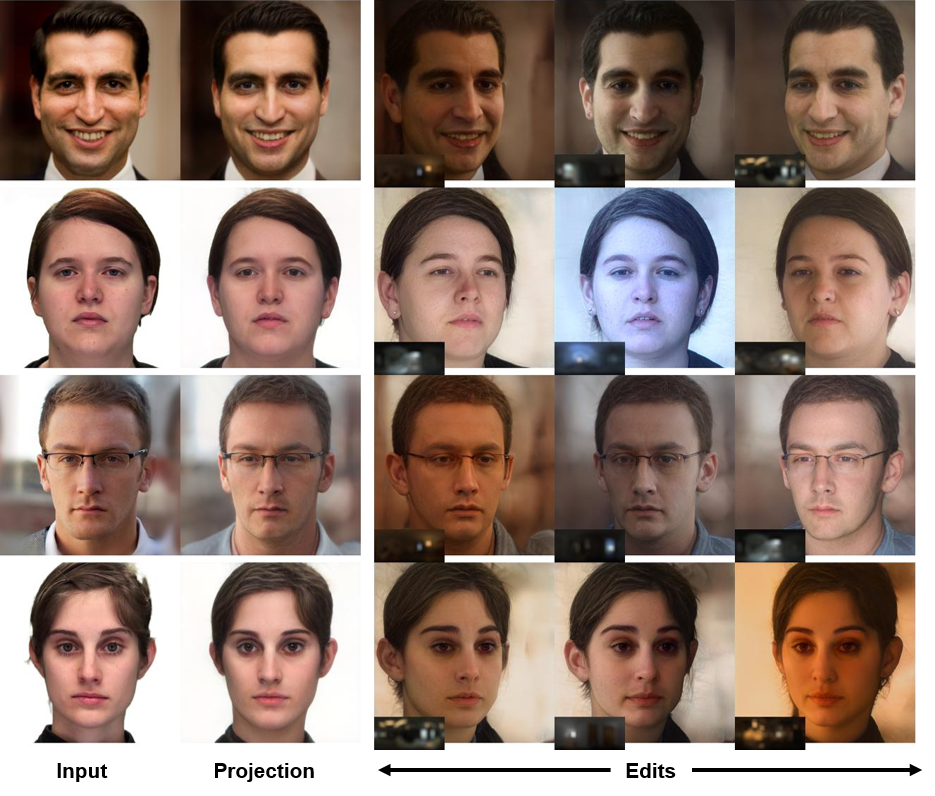}
   \caption{Qualitative illumination and viewpoint editing results. The environment map of the target illumination is shown in the insets. We visualize the StyleGAN projection of the input image (second column). Our method produces photorealistic editing results even under challenging high-frequency light conditions. Portrait images are from \citet{Shih14} (first and third row) and from \citet{Livingstone18} (second and fourth row). Environment maps are from \cite{hold2019deep,gardner2017learning}.}
   \label{fig:quickedits}
\end{figure*} 
\subsection{Discussion}
Existing image-based relighting approaches such as \citet{sun2020light,Zhou_2019_ICCV} rely on much larger trainable networks with several loss functions, such as losses on the input environment map and adversarial losses.  
Approaches for pose editing such as \citet{thies2019deferred,kim2018deep,Siarohin19NeurIPS} rely on conditional generative networks trained with a combination of photometric and adversarial losses. 
Since we rely on a pretrained generator as our backend renderer, our training is much simpler than existing approaches. 
We do not need an adversarial loss as the pretrained generator already synthesises results at a high quality. 
As such, our training is more stable than approaches operating in image space. 
In addition. the StyleGAN latent representation allows for generalisation with high-quality, even when trained on a dataset with as little as 3 identities (Sec.~\ref{sec:generalisation}). 
Many existing methods use specialised network architectures for editing the pose such as landmark-based warping of the features~\cite{Siarohin19NeurIPS}, or rendering of a coarse 3D face model~\cite{kim2018deep,thies2019deferred}. 
Similarly, common relighting networks are designed in a task-specific manner where the illumination is predicted at the bottleneck of the architecture~\cite{sun2020light,Zhou_2019_ICCV}. 
Our design results in a compact and convenient to train PhotoAppNet network that does not require any sophisticated nor specialized network components.
In addition, our method is also faster to train compared to these approaches, since the task we solve is only to transform the latent representation of images, unlike end-to-end approaches which also learn to synthesise high-quality images. 
When trained with 15 identities, our network only takes around 6 hours on a single RTX 8000 GPU to train. 
In contrast, the method of \citet{sun2020light} takes around 26 hours on 4 V100 GPUs for training at the same resolution.

\begin{figure*}
   \centering
   \includegraphics[width=0.92\textwidth]{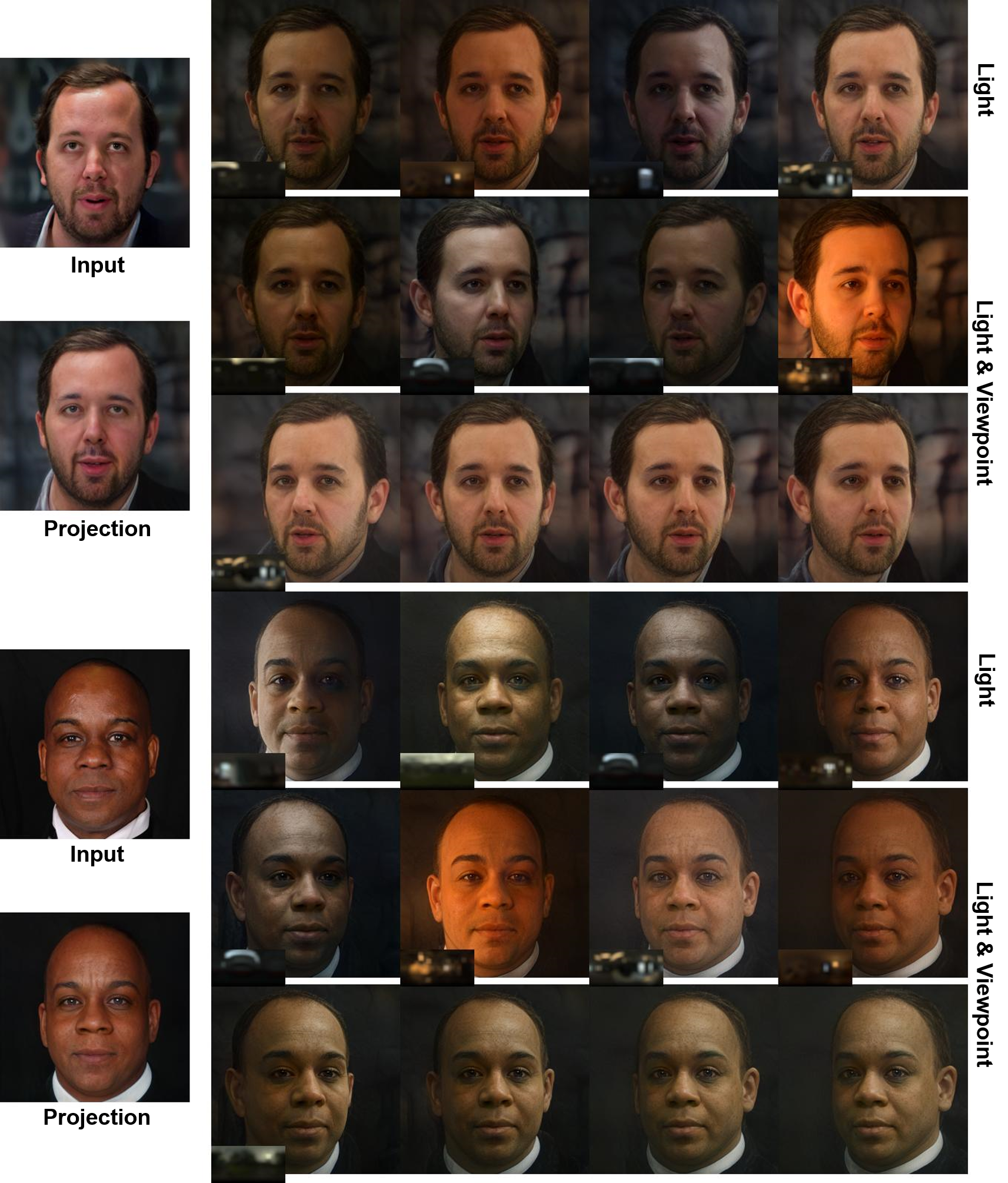}
   \caption{Qualitative illumination and viewpoint editing results. In the first row, we show relighting results where the camera is fixed as in the input. 
   The second row shows results where both illumination and camera pose is edited. 
   The last row shows results with a moving camera under fixed scene illumination. 
   Please note the local shading effects such as shadows, as well as view-dependent effects such as specularities in the image. Portrait images are from \citet{Shih14} (first part) and \citet{stylegan} (second part). Environment maps are from \cite{hold2019deep,gardner2017learning}}
   \label{fig:detailededits2}
\end{figure*} 
\begin{figure*}
   \includegraphics[width=0.92\linewidth]{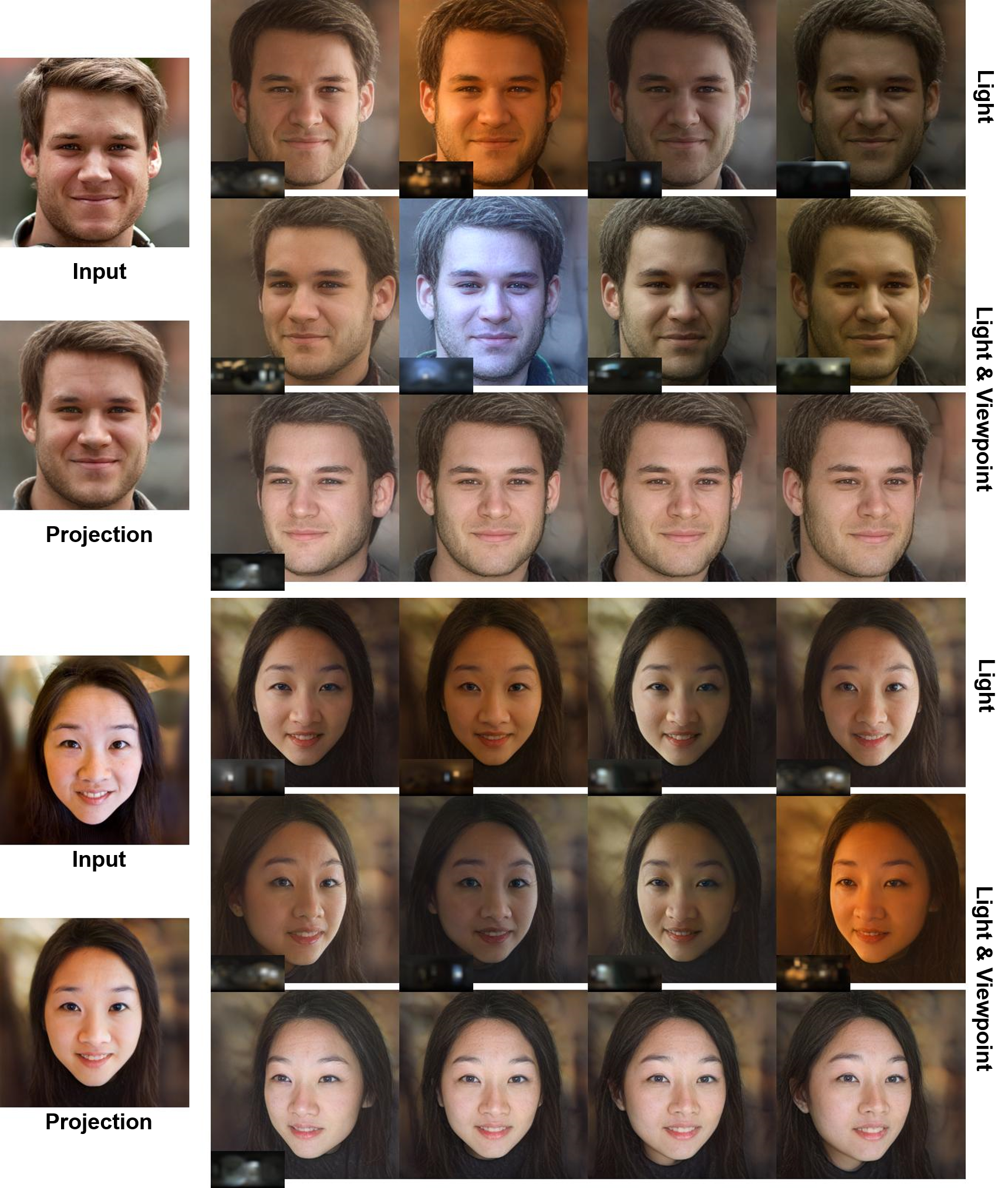}
\caption{Qualitative illumination and viewpoint editing results. In the first row, we show relighting results where the camera is fixed as in the input. 
   The second row shows results where both illumination and camera pose is edited. 
   The last row shows results with a moving camera under fixed scene illumination. 
   Please note the local shading effects such as shadows, as well as view-dependent effects such as specularities in the image. Portrait images are from \citet{Shih14} and environment maps are from \cite{hold2019deep,gardner2017learning}.}   \label{fig:detailededits1}
\end{figure*} 
\section{Results}
\label{Sec:results}

We evaluate our technique both qualitatively and quantitatively on a large set of diverse images. 
The role of the different loss terms is studied in Sec.~\ref{sec:ablation}.  
We compare against several related techniques in Sec.~\ref{sec:comparisons} -- the high-quality relighting approaches of \citet{sipr} and \citet{Zhou_2019_ICCV}, as well as the recent StyleGAN-based image editing approaches of \citet{tewari2020pie} and \citet{abdal2020styleflow} (the latter is concurrent to ours).
Furthermore, we show that our method allows for learning from limited supervised training data by conducting extensive experiments in Sec.~\ref{sec:generalisation}.

\textbf{Data Preparation} We evaluate our approach on portrait images captured in the wild \cite{stylegan,Shih14}. 
All data in our work (including the training data) are cropped and preprocessed as described in \citet{stylegan}.
The images are resized to a resolution of 1024x1024.  
Since we need the ground truth images for quantitative evaluations, we use the test portion of our light-stage dataset composed of images of 41 identities unseen during training. 
We create two test sets, \textit{Set1} has the input and ground truth pairs captured from the same viewpoint while \textit{Set2} includes pairs captured from different viewpoints. 
The HDR environment maps, randomly sampled from the Naval Outdoor and Naval Indoor datasets~\cite{hold2019deep,gardner2017learning} are used to synthesise the pairs with natural illumination conditions. 
Viewpoints are randomly sampled from the 8 cameras of the light-stage setup.
The input and ground truth images are computed using the same environment map in \textit{Set2} for evaluating the viewpoint editing results, while the pairs in \textit{Set1} use different environment maps for relighting evaluations.
\textit{Set1} includes $883$ and \textit{Set2} include $792$ image pairs after finding common set of images which works for all the methods. 
For each pair, we additionally provide a reference image, which is used by related methods to estimate the target illumination and pose in the representation they work with~\cite{sipr,Zhou_2019_ICCV,tewari2020pie,abdal2020styleflow}.
In \textit{Set1}, the reference image is of an identity different from the input identity. 
The ground truth image is directly taken as the reference image for \textit{Set2}, since there can be slight pose variations between different identities for the same camera.
\subsection{High-Fidelty Appearance Editing}

Figs.~\ref{fig:quickedits}, ~\ref{fig:detailededits2}, and~\ref{fig:detailededits1} show simultaneous viewpoint and illumination editing results of our method for various subjects. 
We also show the StyleGAN projection of the input images estimated by \citet{richardson2020encoding}. 
Our approach produces high-quality photorealistic results and synthesises the full portrait, including hair, eyes, mouth, torso and the background, while preserving the identity, expression and other properties (such as facial hair).
Our method works well on people of different races.
Additionally, the results show that our method can preserve a variety of reflectance properties, resulting in effects such as specularities and subsurface scattering. Please note the view-dependent effects such as specularities in the results (nose, forehead...). Our method can synthesise results even under high-frequency light conditions resulting in shadows, even though the StyleGAN network is trained on a dataset of natural images.
In Figs.~\ref{fig:detailededits2}-\ref{fig:detailededits1}, we show more detailed editing results. As it can be noted, the relighting preserve the input pose and identity. Also, our method can change the viewpoint under a fixed environment map (third row for each subject).

\subsection{Ablation Study}
\label{sec:ablation}
\begin{figure*}
   \centering
   \includegraphics[width=1.0\textwidth]{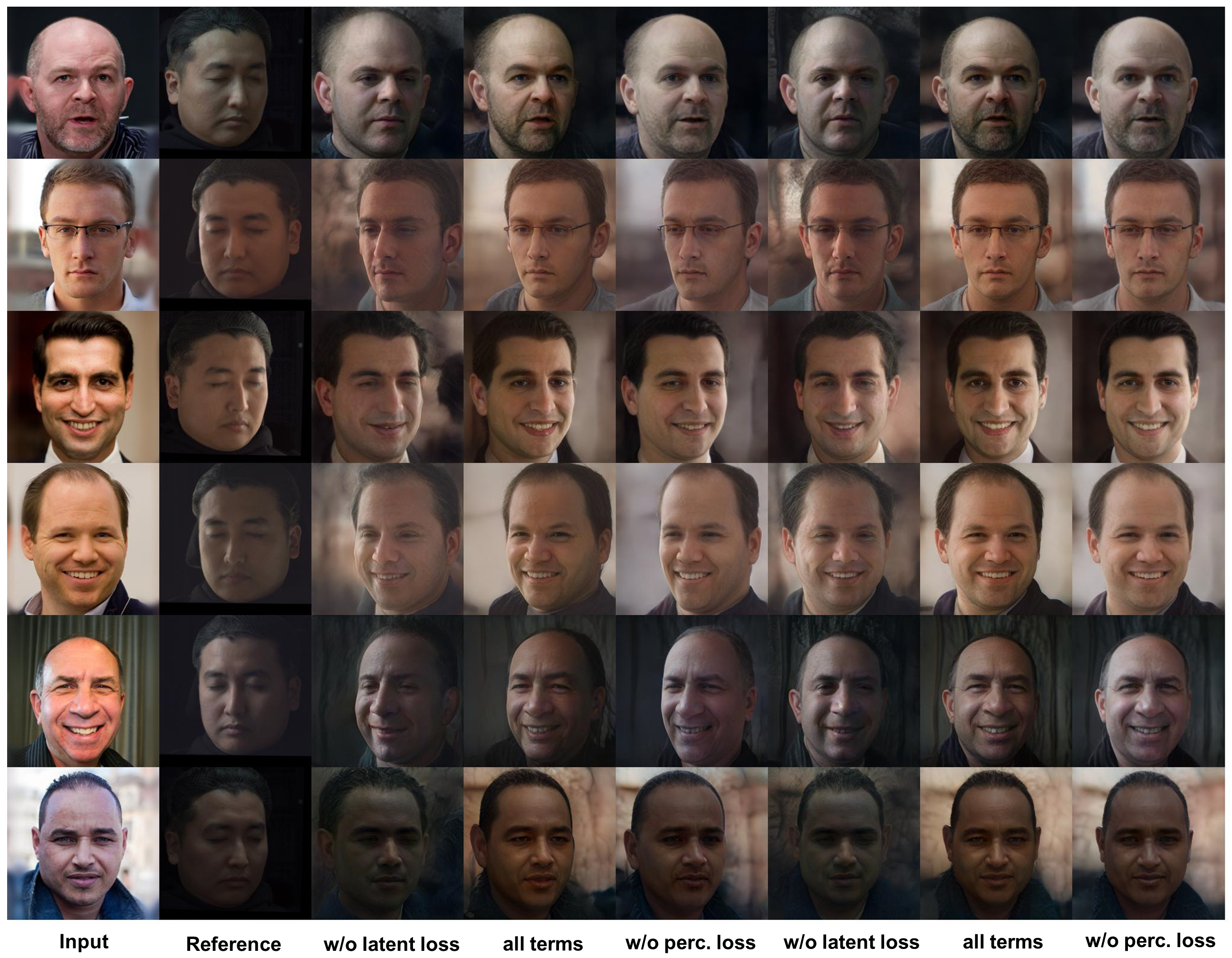}
   \caption{Ablative study on the loss functions. 
   The reference images visualise the target illumination and viewpoint. 
   Removing the latent loss results in distortion of the head geometry and lower quality results. 
   Removing the perceptual term leads to a loss of facial hair and identity preservation such as beards (for e.g., row 1, row 4,5 in light+viewpoint). 
   It also often produces lower-quality results (e.g. row 1,2,6).
   Both terms are necessary for high-quality results.
   Images are from \citet{Shih14} (first column) and \citet{Weyrich2006Analysis} (second column).   
   }
   \label{fig:ablationLosses}
\end{figure*}

In this section, we evaluate the importance of the different loss terms of our objective function (Eq.~\ref{eq:loss}). Results are shown in Fig.~\ref{fig:ablationLosses}.
The target illumination and viewpoint are visualised using a reference image (second column) with the same scene parameters. 
Removing the latent loss leads to clear distortions of the head geometry. 
Only using the perceptual loss leads to results with closed eye expressions, as our training data only consists of people captured with closed eyes. 
We found that the latent loss term helps in generalisation to unseen expressions. 
However, using only the latent loss is not sufficient for high-quality results. In such case, the facial identity and facial hair (see row 1) are not well preserved, and the relighting is not very accurate (see rows 1,2,6). 
A combination of both terms is essential for high-quality.

\begin{figure*}
   \centering
   \includegraphics[width=0.98\textwidth]{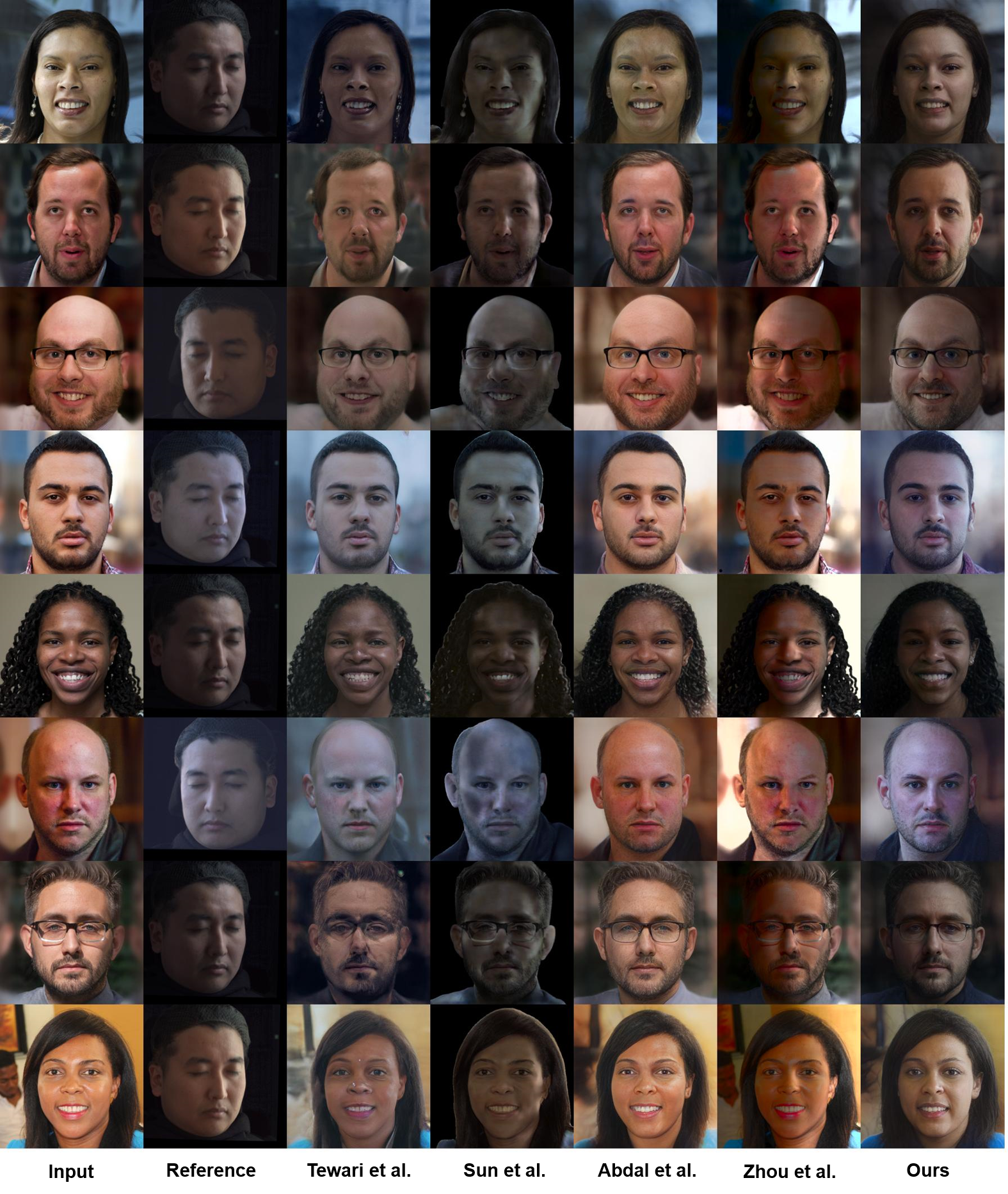}
   \caption{Relighting comparisons. Target illumination is visualised using reference images. 
   Our approach clearly outperforms all existing approaches.
   Here, we compare our method with approaches of ~\citet{tewari2020pie}, ~\citet{sipr}, ~\citet{abdal2020styleflow}, ~\citet{Zhou_2019_ICCV}.
   Images are from \citet{stylegan} (first column, row 1,5,8), \citet{Shih14} (first column, row 2,3,4,6,7) and \citet{Weyrich2006Analysis} (second column).}
   \label{fig:CompLightOnlyReal}
\end{figure*} 
\begin{figure*}
   \centering
   \includegraphics[width=0.65\textwidth]{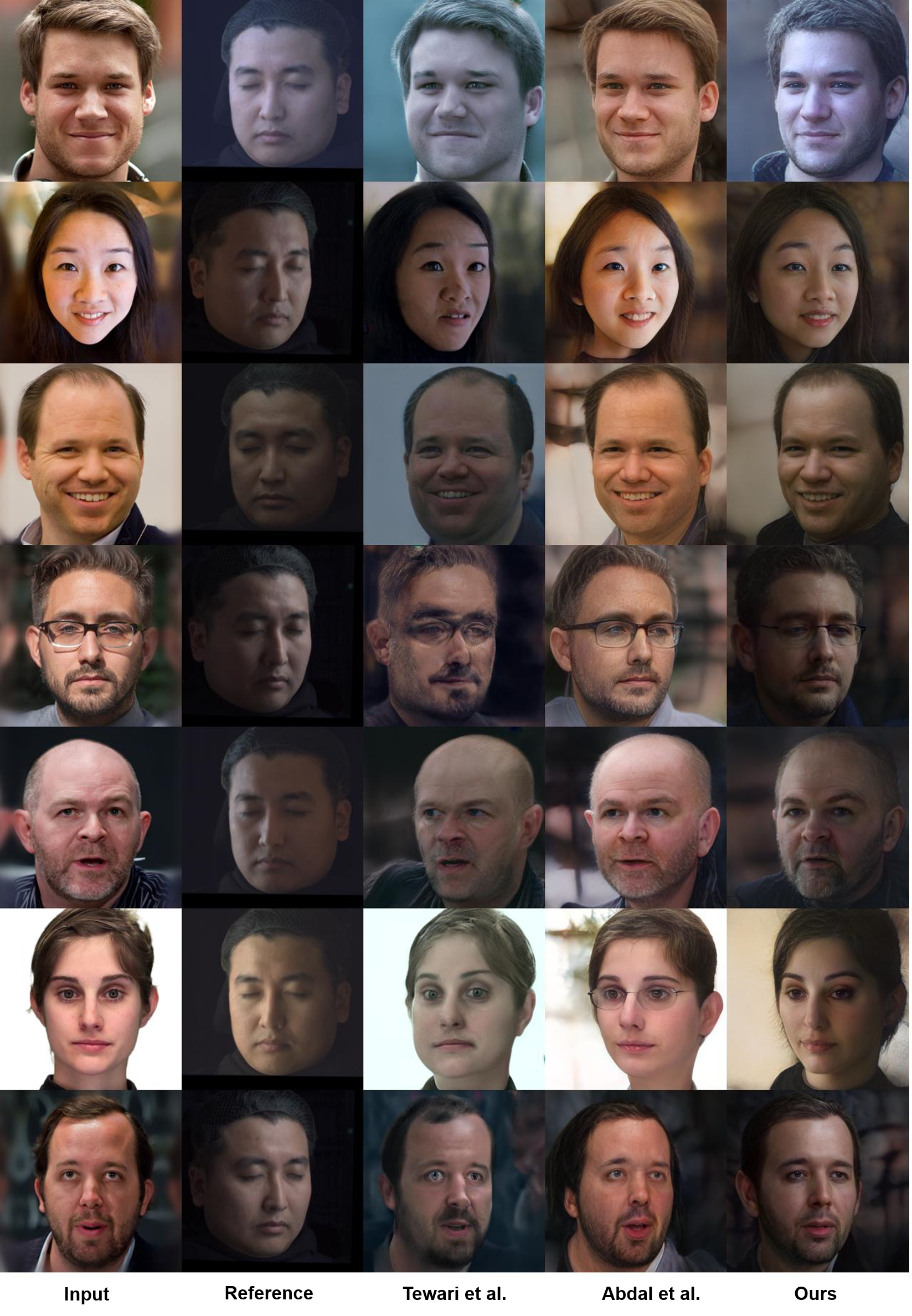}
   \caption{Comparisons to PIE~\cite{tewari2020pie} and StyleFlow~\cite{abdal2020styleflow}. The reference images visualise the target illumination and viewpoint. Our approach produces higher-quality results and clearly outperforms these  methods. Images are from \citet{Shih14} (first column, row 1,2,3,4,5,7), \citet{Livingstone18} (first column, row 6) and \citet{Weyrich2006Analysis} (second column).}
   \label{fig:CompLightPoseReal}
\end{figure*} 
\begin{figure*}
   \centering
   \includegraphics[width=1.15\textwidth, angle=90]{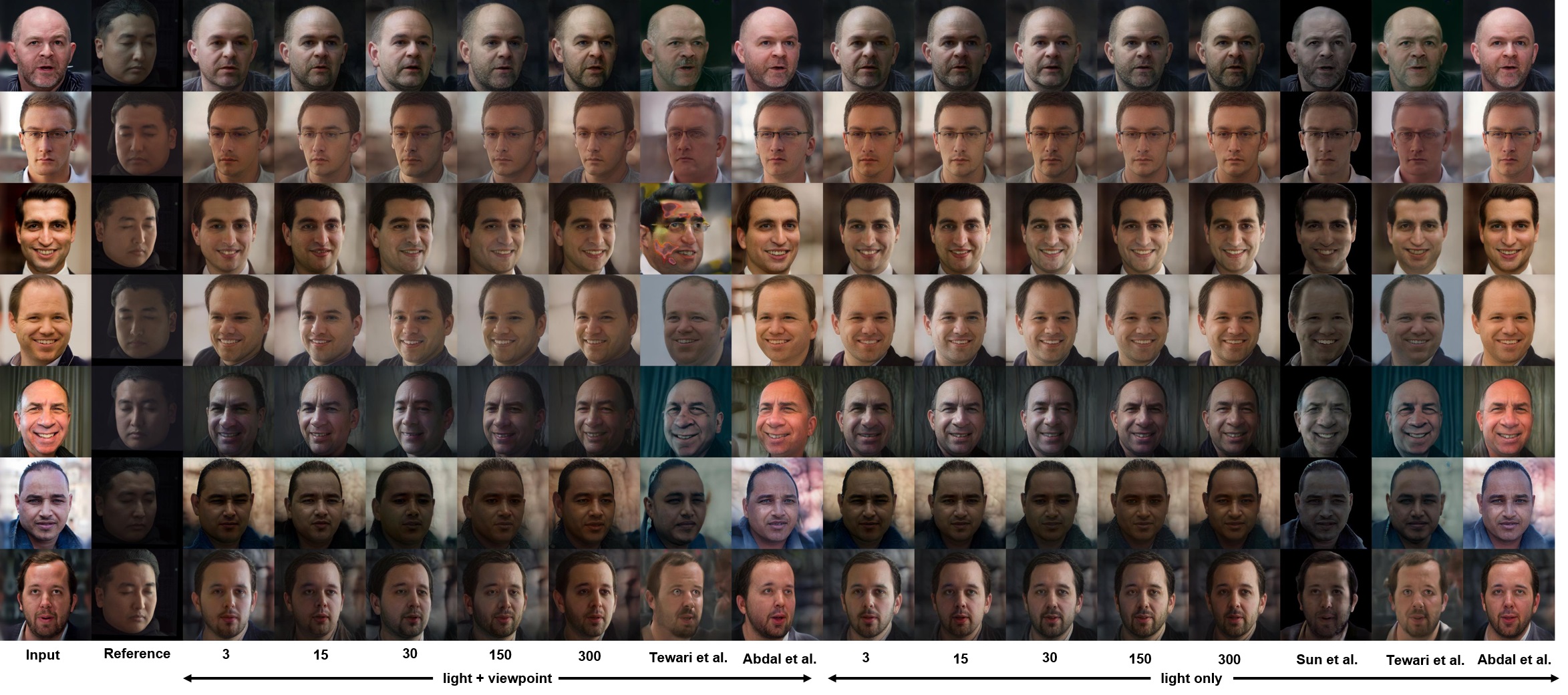}
\caption{Our method allows for training with very limited supervision. 
We show editing results when trained with 3,15,30,150 and 300 identities. 
Our approach produces photorealistic results, and outperforms existing methods even with limited training data. 
Here, we also compare with approaches of ~\citet{sipr}, ~\citet{tewari2020pie} and ~\citet{abdal2020styleflow}.
Images are from \citet{Shih14} (first column) and \citet{Weyrich2006Analysis} (second column).}
   \label{fig:ablationIds}
\end{figure*} 
\begin{figure}
   \centering
   \includegraphics[width=0.45\textwidth]{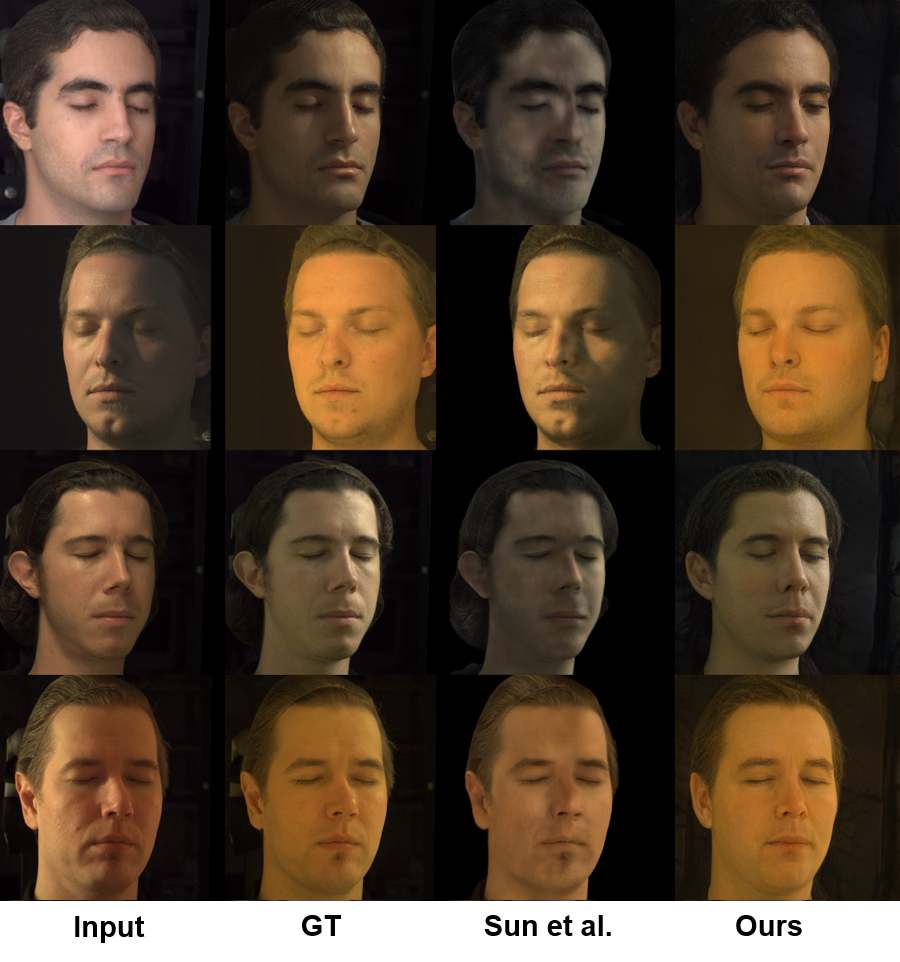}
   \caption{Relighting results on the light stage dataset in comparison with ~\citet{sipr}. Our method obtains higher-quality results which are closer to the ground truth. Images are from \citet{Weyrich2006Analysis}.}
   \label{fig:gtcomp}
\end{figure} 

\subsection{Comparisons to Related Methods}
\label{sec:comparisons}
We compare our method with several state of the art portrait editing approaches. 
We evaluate qualitatively on in the wild data, as well as quantitatively on the test set of the light-stage data. 
We compare with the following approaches: 
\begin{itemize}
    \item The relighting approach of \citet{sipr} which is a data-driven technique trained on a light-stage dataset. It can only edit the scene illumination.
    \item The relighting approach of \citet{Zhou_2019_ICCV} which is trained on synthetic data. It can also only edit the scene illumination. 
    \item PIE~\cite{tewari2020pie} is a method which computes a StyleGAN embedding used to edit the image. It can edit the head pose and scene illumination sequentially (unlike ours, which can perform the edits simultaneously). It is trained without supervised image pairs.
    \item StyleFlow~\cite{abdal2020styleflow}, like PIE can edit images by projecting them onto the StyleGAN latent space. It is also trained without supervised image pairs. Please note that this paper is concurrent to us. However, we provide comparisons for completeness.  
\end{itemize}
We show the relighting comparisons on in the wild data in Fig.~\ref{fig:CompLightOnlyReal}.
Here, the reference image in the second column is used to visualise the target illumination.
Both the light-stage data-driven approach of \citet{sipr} and the synthetic data-driven approach of \citet{Zhou_2019_ICCV}  produce noticeable artifacts.
The approach of \citet{Zhou_2019_ICCV} only uses single channel illumination as input and can thus not capture the overall color tone of the illumination.
The StyleGAN-based approach of \citet{abdal2020styleflow} produces less artifacts, however the quality of relighting is worse than other approaches as it mostly preserves the input lighting. 
In addition, similar to \citet{Zhou_2019_ICCV}, this approach cannot capture the color tone of the environment map. 
PIE~\cite{tewari2020pie} produces better results but it does not capture local illumination effects like our approach (for eg., rows 5,6,7,8) and can produce significant artifacts in some cases (for eg., row 8).
Our approach clearly outperforms all existing methods, demonstrating the effectiveness of a combination of supervised learning and generative modeling. 
It can capture the global color tone as well as local effects such as shadows and specularities. 
It can synthesise the image under harsh lighting (for e.g., rows 7,8) and remove source-lighting related specularities on the glasses (for eg.,  row 5). 
We also compare our method with \citet{sun2020light} on the ground truth light stage images in Fig.~\ref{fig:gtcomp}. 
Our method achieves higher-quality results, closer to the ground truth.

Tab.~\ref{tab:lightcomp} shows the quantitative comparisons with these methods on the light-stage test set (\textit{Set1}). 
We use the Scale invariant-MSE (Si-MSE)~\cite{Zhou_2019_ICCV} and SSIM~\cite{ssim2004} metrics. 
The Si-MSE metric does not penalize global scale offsets between the ground truth and results. 
Our method outperforms all methods using this metric.
The method of \citet{sipr} outperforms other methods on SSIM. 
Since this method uses a U-Net architecture, it is easier to copy the details from the input image, and maintain the pixel correspondences. 
However, visual results clearly show that our approach outperforms all related methods, including that of \citet{sipr} (see Fig.~\ref{fig:CompLightOnlyReal}).  

Fig.~\ref{fig:CompLightPoseReal} shows joint editing of the camera viewpoint and scene illumination for in the wild images. 
The target viewpoint and illumination are visualised using reference images (see second column). 
While PIE~\cite{tewari2020pie} can change the viewpoint, it often distorts the face in an unnatural way (e.g. row 1,2,7). 
It also does not capture local shading effects correctly (e.g. row 1,2,6) and can produce strong artifacts (e.g. row 4). 
PIE solves an optimisation problem to obtain the embedding for each image, which is slow, taking about 10 mins per image.
In contrast, our method is interactive, only requiring 160ms to compute the embedding and edit it. 
StyleFlow~\cite{abdal2020styleflow} can preserve the identity better than PIE, but results in less photorealistic results compared to our method. 
In addition, the relighting results of StyleFlow often fail to capture the input environment map. 
Our approach clearly outperforms both methods both in terms of photorealism as well as the quality of editing. 

Tab.~\ref{tab:lightposecomp} quantitatively compares the joint editing of camera viewpoint and scene illumination.
We use the Si-MSE and SSIM metrics and evaluate on the \textit{Set2} of the light-stage test data. 
Our approach outperforms all methods here in both metrics. 
\subsection{Generalisation with Limited Supervision}
\label{sec:generalisation}
The combination of generative modeling and supervised learning allows us to train from very limited supervised data. 
We show results of training with different number of identities in Fig.~\ref{fig:ablationIds}.
Results of PIE~\cite{tewari2020pie}, StyleFlow~\cite{abdal2020styleflow} and \citet{sun2020light} are also demonstrated. 
Our relighting results outperform related methods both in terms of realism as well as quality of editing, even when trained with as little as 3 identities. 
We also consistently outperform PIE and StyleFlow when editing both viewpoint and illumination, even when trained with 30 identities. 
More identities during training help with better preservation of the facial identity during viewpoint editing. However, very small training data is sufficient for relighting. 

We also quantitatively evaluate these results in Tables~\ref{tab:lightcomp} and~\ref{tab:lightposecomp}.
In both tables, our method outperforms all related approaches using the Si-MSE metric, even when trained with just 3 identities.
All versions of our approach perform similar in terms of SSIM. 
These evaluations show that while larger datasets lead to better results, only limited supervised data is required to outperform the state of the art. 
Finally, despite the limited expressivity of the training dataset (subjects in a single expression with eyes and mouth closed), our method is able to generalise to different expressions, as shown in our results (mouth and eyes open, smiling, etc.) (see Fig.~\ref{fig:dataset}).

\begin{table}[]
\caption{Quantitative comparison with relighting methods. Our approach achieves the lowest Si-MSE numbers. While \citet{sipr} achieves the highest SSIM score, qualitative results show that our method significantly outperforms all existing techniques on in the wild images (see Fig.~\ref{fig:CompLightOnlyReal}).}
\begin{tabular}{@{}llll@{}}
\toprule
 & \begin{tabular}[c]{@{}l@{}} Si-MSE $\downarrow$\end{tabular}
 & \begin{tabular}[c]{@{}l@{}} SSIM $\uparrow$\end{tabular} \\ \midrule
\cite{Zhou_2019_ICCV}  & 0.0037  & 0.9197            \\
 & ($\sigma$= 0.0031) &($\sigma$= 0.0744) \\
\cite{sipr}  & 0.0026 & \textbf{0.9591}  \\
 & ($\sigma$= 0.0024) &($\sigma$= 0.0237) \\
\cite{tewari2020pie}  & 0.0051 & 0.922  \\
 & ($\sigma$= 0.0036) &($\sigma$= 0.029) \\
\cite{abdal2020styleflow}  &0.0082 & 0.8909 \\
 & ($\sigma$=0.0056) &($\sigma$= 0.04) \\
Ours  & \textbf{0.002}  & 0.9199  \\ 
& ($\sigma$=0.001) &($\sigma$= 0.0297) \\
\hdashline
Ours  & \textbf{0.002}  & 0.9192  \\ 
(150 id.) & ($\sigma$=0.001) &($\sigma$= 0.0351) \\
Ours  & \textbf{0.002}  & 0.9188  \\ 
(30 id.) & ($\sigma$=0.001) &($\sigma$= 0.0300) \\
Ours  & \textbf{0.002}  & 0.9191  \\ 
(15 id.) & ($\sigma$=0.001) &($\sigma$= 0.0306) \\
Ours  & \textbf{0.002}  & 0.9193  \\ 
(3 id.) & ($\sigma$=0.001) &($\sigma$= 0.0293) \\
\bottomrule
\end{tabular}
\label{tab:lightcomp}
\end{table}

\begin{table}[]
\caption{Quantitative evaluation with illumination and pose editing methods using \textit{Set2} of the light-stage test set. Our approach outperforms both competing methods, also clearly illustrated using qualitative results (see Fig.~\ref{fig:CompLightPoseReal}).}
\begin{tabular}{@{}llll@{}}
\toprule
 & \begin{tabular}[c]{@{}l@{}} Si-MSE $\downarrow$ \end{tabular}
 & \begin{tabular}[c]{@{}l@{}} SSIM $\uparrow$ \end{tabular} \\ \midrule
\cite{tewari2020pie}   & 0.0067  & 0.9005   \\
 & ($\sigma$= 0.0044) &($\sigma$= 0.0363) \\
\cite{abdal2020styleflow}   & 0.0104  & 0.8812  \\
 & ($\sigma$=0.0071) &($\sigma$= 0.0469) \\
Ours & \textbf{0.0039} & \textbf{0.9086} \\ 
 & ($\sigma$=0.0029) &($\sigma$=0.0307) \\
\hdashline
Ours & \textbf{0.0035} & 0.9050 \\ 
(150 id.) & ($\sigma$=0.0020) &($\sigma$=0.0340) \\
Ours & 0.0040 & 0.9021 \\ 
(30 id.) & ($\sigma$=0.0033) &($\sigma$=0.0312) \\
Ours & 0.0046 & 0.8974 \\ 
(15 id.) & ($\sigma$=0.0027) &($\sigma$=0.0315) \\
Ours & 0.0048 & 0.9000 \\ 
(3 id.) & ($\sigma$=0.0031) &($\sigma$=0.0316) \\
\bottomrule
\end{tabular}
\label{tab:lightposecomp}
\end{table}

\subsection{Preserving the Input Illumination}
\begin{figure}
   \centering
   \includegraphics[width=0.45\textwidth]{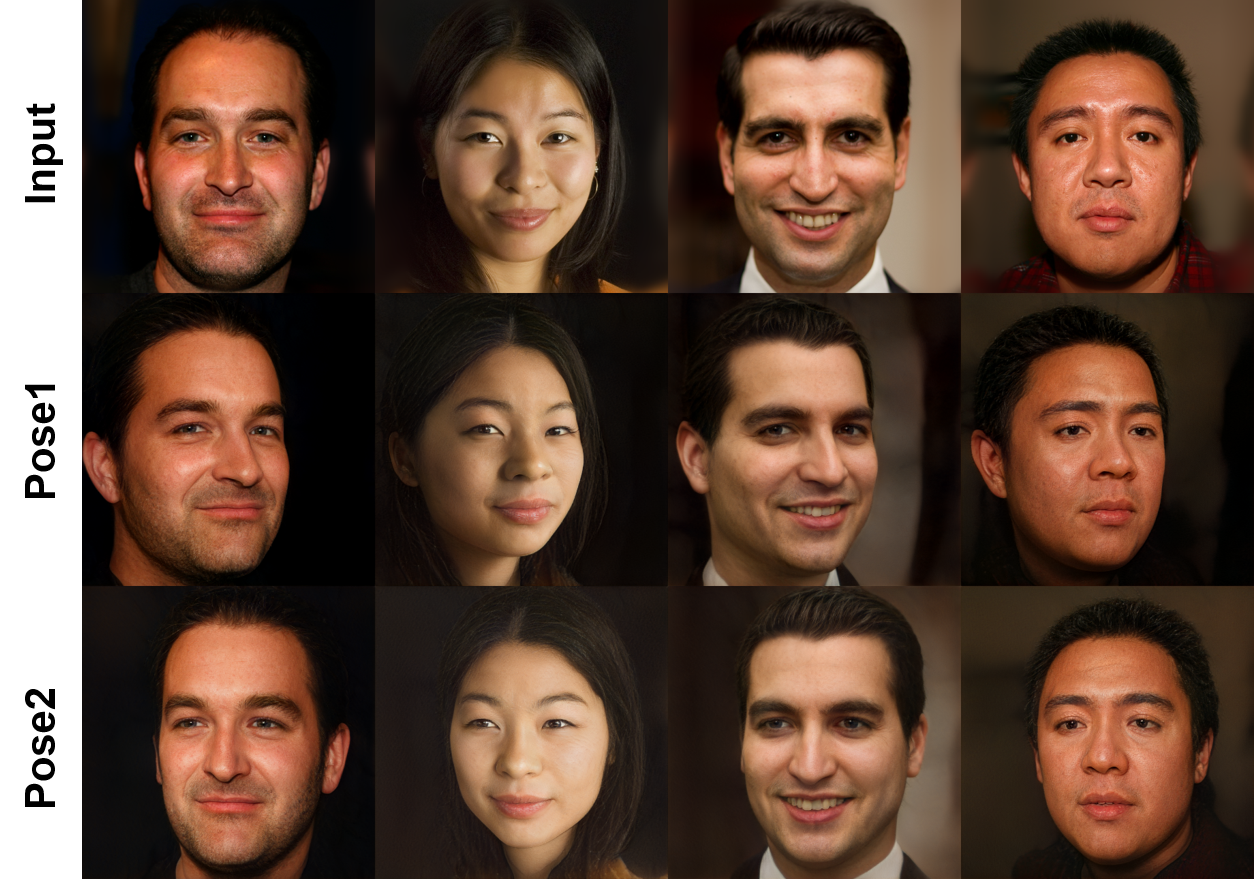}
   \caption{Our method can be easily extended for editing the viewpoint while preserving the input illumination condition. Images are from \citet{Shih14}.}
   \label{fig:viewpointonly}
\end{figure} 
Our method can be easily extended for editing the viewpoint while preserving the input illumination, see Fig.~\ref{fig:viewpointonly}. 
Here, we modify the network architecture in Fig.~\ref{fig:pipeline} by providing 
another binary input similar to $p$, which is set to $0$ when the target illumination is same as the input illumination, and $1$ when they are different.
This design helps in editing both viewpoint and illumination in isolation.

\subsection{Supplemental Material}
In the supplemental video, we show results on videos processed on a per-frame basis. 
We can synthesise the input video from different camera  poses and under different scene illumination while preserving the expressions in the video. 
We also show additional results on a large number of images in the supplemental material.

\section{Limitations}
\begin{figure}
    \centering
    \includegraphics[width=0.48\textwidth]{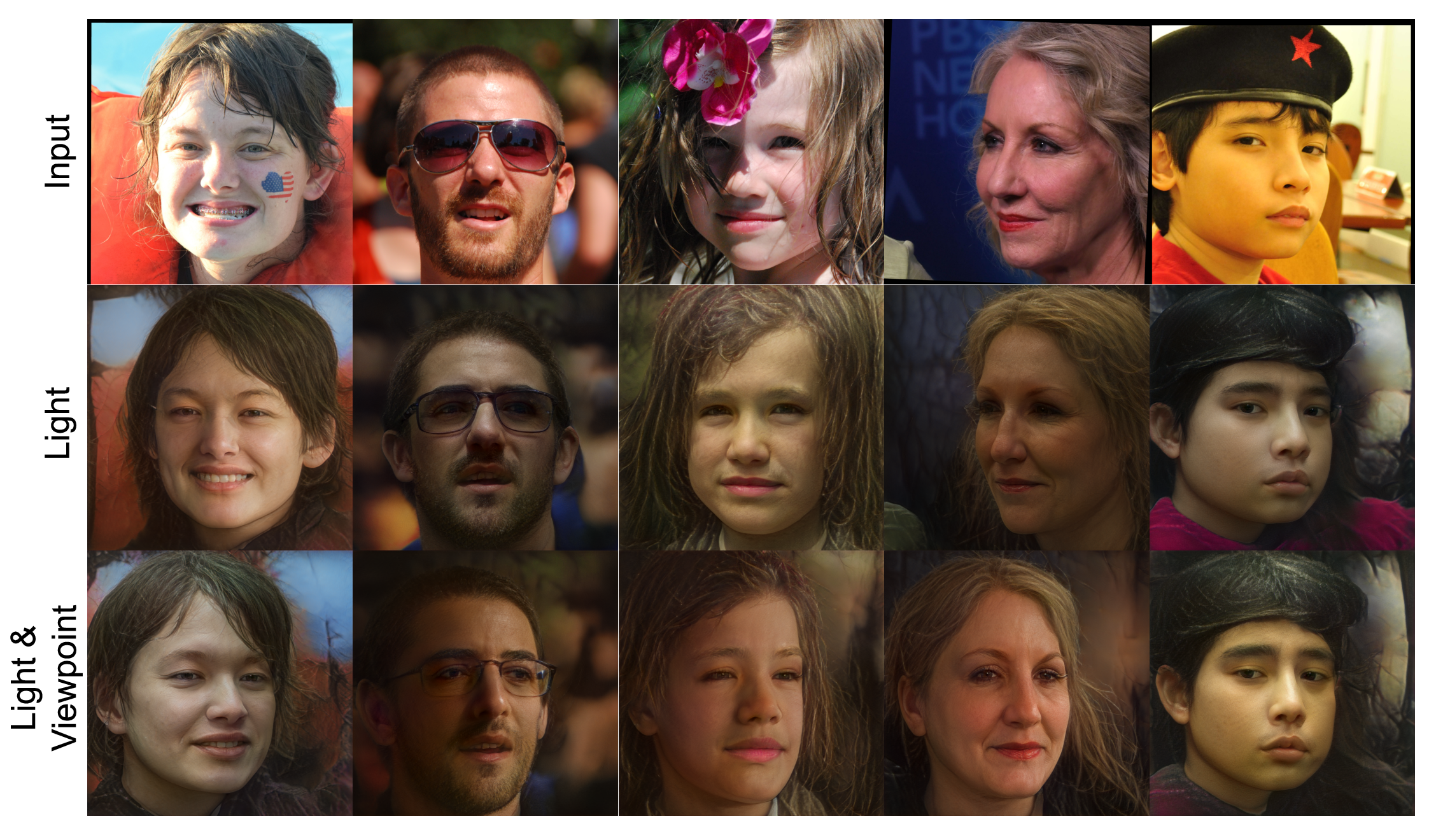}
    \caption{Our method struggles in the presence of accessories such as caps and sunglasses, and background clutter. Extreme pose and illumination editing is also difficult for our method. Images are from \citet{stylegan}.}
    \label{fig:limitations}
\end{figure}
While we demonstrate high-quality results of our approach, several limitations exist, see Fig.~\ref{fig:limitations}. 
Our method can fail to preserve accessories such as caps and glasses in some cases. 
Background clutter can lead to a degradation of quality in the results. 
Our method struggles to preserve the facial identity under large edits for both camera pose and illumination.
While we can preserve the input pose in the results, our method cannot edit the camera viewpoint without changing the illumination. 
In the future, we can use methods that estimate illumination from portrait images~\cite{legendre2020learning} to preserve input illumination when editing the viewpoint.
While we show several high-quality results on video sequences, slight flicker and instability remain. 
A temporal treatment of videos could lead to smoother results. 

\section{Conclusion}
We presented PhotoApp, a method for editing the scene illumination and camera pose in head portraits. 
Our method exploits the advantages of both supervised learning and generative adversarial modeling. 
By designing a supervised learning problem in the latent space of StyleGAN, we achieve high-quality editing results which generalise to in the wild images with significantly more diversity than the training data.  
Through extensive evaluations, we demonstrated that our method outperforms all related techniques, both in terms of realism and editing accuracy.
We further demonstrated that our method can learn from very limited supervised data, achieving high-quality results when trained with as little as 3 identities captured in a single expression. 
While several limitations still exist, we hope that our contributions inspire future work on using generative representations for synthesis applications. 

\begin{acks}
This work was supported by the ERC Consolidator Grant 4DReply (770784). We also acknowledge support from Technicolor and InterDigital. 
We thank Tiancheng Sun for kindly helping us with the comparisons with ~\citet{sipr}. 
\end{acks}

\bibliographystyle{ACM-Reference-Format}
\bibliography{main}


\begin{thebibliography}{43}


\ifx \showCODEN    \undefined \def \showCODEN     #1{\unskip}     \fi
\ifx \showDOI      \undefined \def \showDOI       #1{#1}\fi
\ifx \showISBNx    \undefined \def \showISBNx     #1{\unskip}     \fi
\ifx \showISBNxiii \undefined \def \showISBNxiii  #1{\unskip}     \fi
\ifx \showISSN     \undefined \def \showISSN      #1{\unskip}     \fi
\ifx \showLCCN     \undefined \def \showLCCN      #1{\unskip}     \fi
\ifx \shownote     \undefined \def \shownote      #1{#1}          \fi
\ifx \showarticletitle \undefined \def \showarticletitle #1{#1}   \fi
\ifx \showURL      \undefined \def \showURL       {\relax}        \fi
\providecommand\bibfield[2]{#2}
\providecommand\bibinfo[2]{#2}
\providecommand\natexlab[1]{#1}
\providecommand\showeprint[2][]{arXiv:#2}

\bibitem[\protect\citeauthoryear{Abdal, Zhu, Mitra, and Wonka}{Abdal
  et~al\mbox{.}}{2020}]%
        {abdal2020styleflow}
\bibfield{author}{\bibinfo{person}{Rameen Abdal}, \bibinfo{person}{Peihao Zhu},
  \bibinfo{person}{Niloy Mitra}, {and} \bibinfo{person}{Peter Wonka}.}
  \bibinfo{year}{2020}\natexlab{}.
\newblock \showarticletitle{Styleflow: Attribute-conditioned exploration of
  stylegan-generated images using conditional continuous normalizing flows}.
\newblock \bibinfo{journal}{\emph{arXiv e-prints}} (\bibinfo{year}{2020}),
  \bibinfo{pages}{arXiv--2008}.
\newblock


\bibitem[\protect\citeauthoryear{Averbuch-Elor, Cohen-Or, Kopf, and
  Cohen}{Averbuch-Elor et~al\mbox{.}}{2017}]%
        {Elor17}
\bibfield{author}{\bibinfo{person}{Hadar Averbuch-Elor},
  \bibinfo{person}{Daniel Cohen-Or}, \bibinfo{person}{Johannes Kopf}, {and}
  \bibinfo{person}{Michael~F. Cohen}.} \bibinfo{year}{2017}\natexlab{}.
\newblock \showarticletitle{Bringing Portraits to Life}.
\newblock \bibinfo{journal}{\emph{ACM Trans. on Graph. (Proceedings of SIGGRAPH
  Asia)}} \bibinfo{volume}{36}, \bibinfo{number}{6} (\bibinfo{year}{2017}).
\newblock


\bibitem[\protect\citeauthoryear{B~R, Tewari, Oh, Weyrich, Bickel, Seidel,
  Pfister, Matusik, Elgharib, and Theobalt}{B~R et~al\mbox{.}}{2020}]%
        {mbr_frf}
\bibfield{author}{\bibinfo{person}{Mallikarjun B~R}, \bibinfo{person}{Ayush
  Tewari}, \bibinfo{person}{Tae-Hyun Oh}, \bibinfo{person}{Tim Weyrich},
  \bibinfo{person}{Bernd Bickel}, \bibinfo{person}{Hans-Peter Seidel},
  \bibinfo{person}{Hanspeter Pfister}, \bibinfo{person}{Wojciech Matusik},
  \bibinfo{person}{Mohamed Elgharib}, {and} \bibinfo{person}{Christian
  Theobalt}.} \bibinfo{year}{2020}\natexlab{}.
\newblock \bibinfo{title}{Monocular Reconstruction of Neural Face Reflectance
  Fields}.
\newblock
\newblock
\showeprint[arxiv]{2008.10247}


\bibitem[\protect\citeauthoryear{Collins, Bala, Price, and Susstrunk}{Collins
  et~al\mbox{.}}{2020}]%
        {collins2020editing}
\bibfield{author}{\bibinfo{person}{Edo Collins}, \bibinfo{person}{Raja Bala},
  \bibinfo{person}{Bob Price}, {and} \bibinfo{person}{Sabine Susstrunk}.}
  \bibinfo{year}{2020}\natexlab{}.
\newblock \showarticletitle{Editing in style: Uncovering the local semantics of
  gans}. In \bibinfo{booktitle}{\emph{Proceedings of the IEEE/CVF Conference on
  Computer Vision and Pattern Recognition}}. \bibinfo{pages}{5771--5780}.
\newblock


\bibitem[\protect\citeauthoryear{Debevec, Hawkins, Tchou, Duiker, Sarokin, and
  Sagar}{Debevec et~al\mbox{.}}{2000}]%
        {debevec2000acquiring}
\bibfield{author}{\bibinfo{person}{Paul Debevec}, \bibinfo{person}{Tim
  Hawkins}, \bibinfo{person}{Chris Tchou}, \bibinfo{person}{Haarm-Pieter
  Duiker}, \bibinfo{person}{Westley Sarokin}, {and} \bibinfo{person}{Mark
  Sagar}.} \bibinfo{year}{2000}\natexlab{}.
\newblock \showarticletitle{Acquiring the reflectance field of a human face}.
  In \bibinfo{booktitle}{\emph{Annual conference on Computer graphics and
  interactive techniques}}.
\newblock


\bibitem[\protect\citeauthoryear{Gardner, Sunkavalli, Yumer, Shen, Gambaretto,
  Gagn{\'e}, and Lalonde}{Gardner et~al\mbox{.}}{2017}]%
        {gardner2017learning}
\bibfield{author}{\bibinfo{person}{Marc-Andr{\'e} Gardner},
  \bibinfo{person}{Kalyan Sunkavalli}, \bibinfo{person}{Ersin Yumer},
  \bibinfo{person}{Xiaohui Shen}, \bibinfo{person}{Emiliano Gambaretto},
  \bibinfo{person}{Christian Gagn{\'e}}, {and}
  \bibinfo{person}{Jean-Fran{\c{c}}ois Lalonde}.}
  \bibinfo{year}{2017}\natexlab{}.
\newblock \showarticletitle{Learning to predict indoor illumination from a
  single image}.
\newblock \bibinfo{journal}{\emph{ACM Trans. on Graph. (Proceedings of SIGGRAPH
  Asia)}} \bibinfo{volume}{36}, \bibinfo{number}{6}, Article
  \bibinfo{articleno}{176} (\bibinfo{year}{2017}).
\newblock


\bibitem[\protect\citeauthoryear{Geng, Shao, Zheng, Weng, and Zhou}{Geng
  et~al\mbox{.}}{2018}]%
        {Geng18}
\bibfield{author}{\bibinfo{person}{Jiahao Geng}, \bibinfo{person}{Tianjia
  Shao}, \bibinfo{person}{Youyi Zheng}, \bibinfo{person}{Yanlin Weng}, {and}
  \bibinfo{person}{Kun Zhou}.} \bibinfo{year}{2018}\natexlab{}.
\newblock \showarticletitle{Warp-Guided GANs for Single-Photo Facial
  Animation}.
\newblock \bibinfo{journal}{\emph{ACM Trans. on Graph. (Proceedings of SIGGRAPH
  Asia)}} \bibinfo{volume}{37}, \bibinfo{number}{6} (\bibinfo{year}{2018}).
\newblock


\bibitem[\protect\citeauthoryear{Ghosh, Fyffe, Tunwattanapong, Busch, Yu, and
  Debevec}{Ghosh et~al\mbox{.}}{2011}]%
        {mvfc_ghosh}
\bibfield{author}{\bibinfo{person}{Abhijeet Ghosh}, \bibinfo{person}{Graham
  Fyffe}, \bibinfo{person}{Borom Tunwattanapong}, \bibinfo{person}{Jay Busch},
  \bibinfo{person}{Xueming Yu}, {and} \bibinfo{person}{Paul Debevec}.}
  \bibinfo{year}{2011}\natexlab{}.
\newblock \showarticletitle{Multiview Face Capture Using Polarized Spherical
  Gradient Illumination}.
\newblock \bibinfo{journal}{\emph{ACM Trans. on Graph.}} \bibinfo{volume}{30},
  \bibinfo{number}{6} (\bibinfo{date}{Dec.} \bibinfo{year}{2011}),
  \bibinfo{pages}{1–10}.
\newblock


\bibitem[\protect\citeauthoryear{H{\"a}rk{\"o}nen, Hertzmann, Lehtinen, and
  Paris}{H{\"a}rk{\"o}nen et~al\mbox{.}}{2020}]%
        {harkonen2020ganspace}
\bibfield{author}{\bibinfo{person}{Erik H{\"a}rk{\"o}nen},
  \bibinfo{person}{Aaron Hertzmann}, \bibinfo{person}{Jaakko Lehtinen}, {and}
  \bibinfo{person}{Sylvain Paris}.} \bibinfo{year}{2020}\natexlab{}.
\newblock \showarticletitle{Ganspace: Discovering interpretable gan controls}.
\newblock \bibinfo{journal}{\emph{arXiv preprint arXiv:2004.02546}}
  (\bibinfo{year}{2020}).
\newblock


\bibitem[\protect\citeauthoryear{Hold-Geoffroy, Athawale, and
  Lalonde}{Hold-Geoffroy et~al\mbox{.}}{2019}]%
        {hold2019deep}
\bibfield{author}{\bibinfo{person}{Yannick Hold-Geoffroy},
  \bibinfo{person}{Akshaya Athawale}, {and}
  \bibinfo{person}{Jean-Fran{\c{c}}ois Lalonde}.}
  \bibinfo{year}{2019}\natexlab{}.
\newblock \showarticletitle{Deep sky modeling for single image outdoor lighting
  estimation}. In \bibinfo{booktitle}{\emph{Computer Vision and Pattern
  Recognition (CVPR)}}.
\newblock


\bibitem[\protect\citeauthoryear{Jing and Tian}{Jing and Tian}{2020}]%
        {jing2020self}
\bibfield{author}{\bibinfo{person}{Longlong Jing} {and} \bibinfo{person}{Yingli
  Tian}.} \bibinfo{year}{2020}\natexlab{}.
\newblock \showarticletitle{Self-supervised visual feature learning with deep
  neural networks: A survey}.
\newblock \bibinfo{journal}{\emph{IEEE Transactions on Pattern Analysis and
  Machine Intelligence}} (\bibinfo{year}{2020}).
\newblock


\bibitem[\protect\citeauthoryear{Kajiya}{Kajiya}{1986}]%
        {Kajiya86}
\bibfield{author}{\bibinfo{person}{James~T. Kajiya}.}
  \bibinfo{year}{1986}\natexlab{}.
\newblock \showarticletitle{The Rendering Equation}.
\newblock \bibinfo{journal}{\emph{SIGGRAPH Computer Graphics}}
  \bibinfo{volume}{20}, \bibinfo{number}{4} (\bibinfo{year}{1986}),
  \bibinfo{pages}{143–150}.
\newblock
\showISSN{0097-8930}
\urldef\tempurl%
\url{https://doi.org/10.1145/15886.15902}
\showURL{%
\tempurl}


\bibitem[\protect\citeauthoryear{{Karras}, {Laine}, and {Aila}}{{Karras}
  et~al\mbox{.}}{2019}]%
        {stylegan}
\bibfield{author}{\bibinfo{person}{T. {Karras}}, \bibinfo{person}{S. {Laine}},
  {and} \bibinfo{person}{T. {Aila}}.} \bibinfo{year}{2019}\natexlab{}.
\newblock \showarticletitle{A Style-Based Generator Architecture for Generative
  Adversarial Networks}. In \bibinfo{booktitle}{\emph{Computer Vision and
  Pattern Recognition (CVPR)}}.
\newblock


\bibitem[\protect\citeauthoryear{Karras, Laine, Aittala, Hellsten, Lehtinen,
  and Aila}{Karras et~al\mbox{.}}{2020}]%
        {stylegan2}
\bibfield{author}{\bibinfo{person}{Tero Karras}, \bibinfo{person}{Samuli
  Laine}, \bibinfo{person}{Miika Aittala}, \bibinfo{person}{Janne Hellsten},
  \bibinfo{person}{Jaakko Lehtinen}, {and} \bibinfo{person}{Timo Aila}.}
  \bibinfo{year}{2020}\natexlab{}.
\newblock \showarticletitle{Analyzing and Improving the Image Quality of
  {StyleGAN}}. In \bibinfo{booktitle}{\emph{Computer Vision and Pattern
  Recognition (CVPR)}}.
\newblock


\bibitem[\protect\citeauthoryear{Kim, Garrido, Tewari, Xu, Thies, Nie{\ss}ner,
  P{\'e}rez, Richardt, Zoll{\"o}fer, and Theobalt}{Kim et~al\mbox{.}}{2018}]%
        {kim2018deep}
\bibfield{author}{\bibinfo{person}{Hyeongwoo Kim}, \bibinfo{person}{Pablo
  Garrido}, \bibinfo{person}{Ayush Tewari}, \bibinfo{person}{Weipeng Xu},
  \bibinfo{person}{Justus Thies}, \bibinfo{person}{Matthias Nie{\ss}ner},
  \bibinfo{person}{Patrick P{\'e}rez}, \bibinfo{person}{Christian Richardt},
  \bibinfo{person}{Michael Zoll{\"o}fer}, {and} \bibinfo{person}{Christian
  Theobalt}.} \bibinfo{year}{2018}\natexlab{}.
\newblock \showarticletitle{Deep Video Portraits}.
\newblock \bibinfo{journal}{\emph{ACM Trans. on Graph. (Proceedings of
  SIGGRAPH)}} \bibinfo{volume}{37}, \bibinfo{number}{4} (\bibinfo{year}{2018}),
  \bibinfo{pages}{163}.
\newblock


\bibitem[\protect\citeauthoryear{Krizhevsky, Sutskever, and Hinton}{Krizhevsky
  et~al\mbox{.}}{2012}]%
        {krizhevsky2012imagenet}
\bibfield{author}{\bibinfo{person}{Alex Krizhevsky}, \bibinfo{person}{Ilya
  Sutskever}, {and} \bibinfo{person}{Geoffrey~E Hinton}.}
  \bibinfo{year}{2012}\natexlab{}.
\newblock \showarticletitle{Imagenet classification with deep convolutional
  neural networks}.
\newblock \bibinfo{journal}{\emph{Advances in neural information processing
  systems}}  \bibinfo{volume}{25} (\bibinfo{year}{2012}),
  \bibinfo{pages}{1097--1105}.
\newblock


\bibitem[\protect\citeauthoryear{Lattas, Moschoglou, Gecer, Ploumpis,
  Triantafyllou, Ghosh, and Zafeiriou}{Lattas et~al\mbox{.}}{2020}]%
        {lattas2020avatarme}
\bibfield{author}{\bibinfo{person}{Alexandros Lattas},
  \bibinfo{person}{Stylianos Moschoglou}, \bibinfo{person}{Baris Gecer},
  \bibinfo{person}{Stylianos Ploumpis}, \bibinfo{person}{Vasileios
  Triantafyllou}, \bibinfo{person}{Abhijeet Ghosh}, {and}
  \bibinfo{person}{Stefanos Zafeiriou}.} \bibinfo{year}{2020}\natexlab{}.
\newblock \showarticletitle{AvatarMe: Realistically Renderable 3D Facial
  Reconstruction "in-the-wild"}. In \bibinfo{booktitle}{\emph{Computer Vision
  and Pattern Recognition (CVPR)}}.
\newblock


\bibitem[\protect\citeauthoryear{LeGendre, Ma, Pandey, Fanello, Rhemann,
  Dourgarian, Busch, and Debevec}{LeGendre et~al\mbox{.}}{2020}]%
        {legendre2020learning}
\bibfield{author}{\bibinfo{person}{Chloe LeGendre}, \bibinfo{person}{Wan-Chun
  Ma}, \bibinfo{person}{Rohit Pandey}, \bibinfo{person}{Sean Fanello},
  \bibinfo{person}{Christoph Rhemann}, \bibinfo{person}{Jason Dourgarian},
  \bibinfo{person}{Jay Busch}, {and} \bibinfo{person}{Paul Debevec}.}
  \bibinfo{year}{2020}\natexlab{}.
\newblock \showarticletitle{Learning Illumination from Diverse Portraits}.
\newblock In \bibinfo{booktitle}{\emph{SIGGRAPH Asia 2020 Technical
  Communications}}. \bibinfo{pages}{1--4}.
\newblock


\bibitem[\protect\citeauthoryear{Livingstone and Russo}{Livingstone and
  Russo}{2018}]%
        {Livingstone18}
\bibfield{author}{\bibinfo{person}{Steven~R. Livingstone} {and}
  \bibinfo{person}{Frank~A. Russo}.} \bibinfo{year}{2018}\natexlab{}.
\newblock \bibinfo{booktitle}{\emph{{The Ryerson Audio-Visual Database of
  Emotional Speech and Song (RAVDESS)}}}.
\newblock
\urldef\tempurl%
\url{https://doi.org/10.5281/zenodo.1188976}
\showDOI{\tempurl}
\newblock
\shownote{{Funding Information Natural Sciences and Engineering Research
  Council of Canada: 2012-341583 Hear the world research chair in music and
  emotional speech from Phonak}.}


\bibitem[\protect\citeauthoryear{Meka, H\"{a}ne, Pandey, Zollh\"{o}fer,
  Fanello, Fyffe, Kowdle, Yu, Busch, Dourgarian, Denny, Bouaziz, Lincoln,
  Whalen, Harvey, Taylor, Izadi, Tagliasacchi, Debevec, Theobalt, Valentin, and
  Rhemann}{Meka et~al\mbox{.}}{2019}]%
        {Meka19}
\bibfield{author}{\bibinfo{person}{Abhimitra Meka}, \bibinfo{person}{Christian
  H\"{a}ne}, \bibinfo{person}{Rohit Pandey}, \bibinfo{person}{Michael
  Zollh\"{o}fer}, \bibinfo{person}{Sean Fanello}, \bibinfo{person}{Graham
  Fyffe}, \bibinfo{person}{Adarsh Kowdle}, \bibinfo{person}{Xueming Yu},
  \bibinfo{person}{Jay Busch}, \bibinfo{person}{Jason Dourgarian},
  \bibinfo{person}{Peter Denny}, \bibinfo{person}{Sofien Bouaziz},
  \bibinfo{person}{Peter Lincoln}, \bibinfo{person}{Matt Whalen},
  \bibinfo{person}{Geoff Harvey}, \bibinfo{person}{Jonathan Taylor},
  \bibinfo{person}{Shahram Izadi}, \bibinfo{person}{Andrea Tagliasacchi},
  \bibinfo{person}{Paul Debevec}, \bibinfo{person}{Christian Theobalt},
  \bibinfo{person}{Julien Valentin}, {and} \bibinfo{person}{Christoph
  Rhemann}.} \bibinfo{year}{2019}\natexlab{}.
\newblock \showarticletitle{Deep Reflectance Fields: High-Quality Facial
  Reflectance Field Inference from Color Gradient Illumination}.
\newblock \bibinfo{journal}{\emph{ACM Trans. on Graph. (Proceedings of
  SIGGRAPH)}} \bibinfo{volume}{38}, \bibinfo{number}{4} (\bibinfo{year}{2019}).
\newblock


\bibitem[\protect\citeauthoryear{Nagano, Seo, Xing, Wei, Li, Saito, Agarwal,
  Fursund, and Li}{Nagano et~al\mbox{.}}{2018}]%
        {nagano2018pagan}
\bibfield{author}{\bibinfo{person}{Koki Nagano}, \bibinfo{person}{Jaewoo Seo},
  \bibinfo{person}{Jun Xing}, \bibinfo{person}{Lingyu Wei},
  \bibinfo{person}{Zimo Li}, \bibinfo{person}{Shunsuke Saito},
  \bibinfo{person}{Aviral Agarwal}, \bibinfo{person}{Jens Fursund}, {and}
  \bibinfo{person}{Hao Li}.} \bibinfo{year}{2018}\natexlab{}.
\newblock \showarticletitle{paGAN: real-time avatars using dynamic textures}.
  In \bibinfo{booktitle}{\emph{ACM Trans. on Graph. (Proceedings of SIGGRAPH
  Asia)}}. \bibinfo{pages}{258}.
\newblock


\bibitem[\protect\citeauthoryear{Nestmeyer, Lalonde, Matthews, and
  Lehrmann}{Nestmeyer et~al\mbox{.}}{2020}]%
        {nestmeyer2020faceRelighting}
\bibfield{author}{\bibinfo{person}{Thomas Nestmeyer},
  \bibinfo{person}{Jean-François Lalonde}, \bibinfo{person}{Iain Matthews},
  {and} \bibinfo{person}{Andreas~M Lehrmann}.} \bibinfo{year}{2020}\natexlab{}.
\newblock \showarticletitle{Learning Physics-guided Face Relighting under
  Directional Light}. In \bibinfo{booktitle}{\emph{Computer Vision and Pattern
  Recognition (CVPR)}}.
\newblock


\bibitem[\protect\citeauthoryear{Richardson, Alaluf, Patashnik, Nitzan, Azar,
  Shapiro, and Cohen-Or}{Richardson et~al\mbox{.}}{2020}]%
        {richardson2020encoding}
\bibfield{author}{\bibinfo{person}{Elad Richardson}, \bibinfo{person}{Yuval
  Alaluf}, \bibinfo{person}{Or Patashnik}, \bibinfo{person}{Yotam Nitzan},
  \bibinfo{person}{Yaniv Azar}, \bibinfo{person}{Stav Shapiro}, {and}
  \bibinfo{person}{Daniel Cohen-Or}.} \bibinfo{year}{2020}\natexlab{}.
\newblock \showarticletitle{Encoding in Style: a StyleGAN Encoder for
  Image-to-Image Translation}.
\newblock \bibinfo{journal}{\emph{arXiv preprint arXiv:2008.00951}}
  (\bibinfo{year}{2020}).
\newblock


\bibitem[\protect\citeauthoryear{Sengupta, Kanazawa, Castillo, and
  Jacobs}{Sengupta et~al\mbox{.}}{2018}]%
        {sfsnetSengupta18}
\bibfield{author}{\bibinfo{person}{Soumyadip Sengupta}, \bibinfo{person}{Angjoo
  Kanazawa}, \bibinfo{person}{Carlos~D. Castillo}, {and}
  \bibinfo{person}{David~W. Jacobs}.} \bibinfo{year}{2018}\natexlab{}.
\newblock \showarticletitle{SfSNet: Learning Shape, Refectance and Illuminance
  of Faces in the Wild}. In \bibinfo{booktitle}{\emph{Computer Vision and
  Pattern Regognition (CVPR)}}.
\newblock


\bibitem[\protect\citeauthoryear{Shen, Gu, Tang, and Zhou}{Shen
  et~al\mbox{.}}{2020}]%
        {shen2020interpreting}
\bibfield{author}{\bibinfo{person}{Yujun Shen}, \bibinfo{person}{Jinjin Gu},
  \bibinfo{person}{Xiaoou Tang}, {and} \bibinfo{person}{Bolei Zhou}.}
  \bibinfo{year}{2020}\natexlab{}.
\newblock \showarticletitle{Interpreting the Latent Space of GANs for Semantic
  Face Editing}. In \bibinfo{booktitle}{\emph{CVPR}}.
\newblock


\bibitem[\protect\citeauthoryear{Shih, Paris, Barnes, Freeman, and Durand}{Shih
  et~al\mbox{.}}{2014}]%
        {Shih14}
\bibfield{author}{\bibinfo{person}{YiChang Shih}, \bibinfo{person}{Sylvain
  Paris}, \bibinfo{person}{Connelly Barnes}, \bibinfo{person}{William~T.
  Freeman}, {and} \bibinfo{person}{Fr\'{e}do Durand}.}
  \bibinfo{year}{2014}\natexlab{}.
\newblock \showarticletitle{Style Transfer for Headshot Portraits}.
\newblock \bibinfo{journal}{\emph{ACM Trans. on Graph. (Proceedings of
  SIGGRAPH)}} \bibinfo{volume}{33}, \bibinfo{number}{4}, Article
  \bibinfo{articleno}{148} (\bibinfo{year}{2014}),
  \bibinfo{numpages}{14}~pages.
\newblock


\bibitem[\protect\citeauthoryear{{Shu}, {Yumer}, {Hadap}, {Sunkavalli},
  {Shechtman}, and {Samaras}}{{Shu} et~al\mbox{.}}{2017}]%
        {Shu17NeuralEditing}
\bibfield{author}{\bibinfo{person}{Z. {Shu}}, \bibinfo{person}{E. {Yumer}},
  \bibinfo{person}{S. {Hadap}}, \bibinfo{person}{K. {Sunkavalli}},
  \bibinfo{person}{E. {Shechtman}}, {and} \bibinfo{person}{D. {Samaras}}.}
  \bibinfo{year}{2017}\natexlab{}.
\newblock \showarticletitle{Neural Face Editing with Intrinsic Image
  Disentangling}. In \bibinfo{booktitle}{\emph{Computer Vision and Pattern
  Recognition (CVPR)}}. \bibinfo{pages}{5444--5453}.
\newblock


\bibitem[\protect\citeauthoryear{Siarohin, Lathuilière, Tulyakov, Ricci, and
  Sebe}{Siarohin et~al\mbox{.}}{2019}]%
        {Siarohin19NeurIPS}
\bibfield{author}{\bibinfo{person}{Aliaksandr Siarohin},
  \bibinfo{person}{Stéphane Lathuilière}, \bibinfo{person}{Sergey Tulyakov},
  \bibinfo{person}{Elisa Ricci}, {and} \bibinfo{person}{Nicu Sebe}.}
  \bibinfo{year}{2019}\natexlab{}.
\newblock \showarticletitle{First Order Motion Model for Image Animation}. In
  \bibinfo{booktitle}{\emph{Conference on Neural Information Processing Systems
  (NeurIPS)}}.
\newblock


\bibitem[\protect\citeauthoryear{Sun, Barron, Tsai, Xu, Yu, Fyffe, Rhemann,
  Busch, Debevec, and Ramamoorthi}{Sun et~al\mbox{.}}{2019}]%
        {sipr}
\bibfield{author}{\bibinfo{person}{Tiancheng Sun}, \bibinfo{person}{Jonathan~T.
  Barron}, \bibinfo{person}{Yun-Ta Tsai}, \bibinfo{person}{Zexiang Xu},
  \bibinfo{person}{Xueming Yu}, \bibinfo{person}{Graham Fyffe},
  \bibinfo{person}{Christoph Rhemann}, \bibinfo{person}{Jay Busch},
  \bibinfo{person}{Paul Debevec}, {and} \bibinfo{person}{Ravi Ramamoorthi}.}
  \bibinfo{year}{2019}\natexlab{}.
\newblock \showarticletitle{Single Image Portrait Relighting}.
\newblock  \bibinfo{volume}{38}, \bibinfo{number}{4}, Article
  \bibinfo{articleno}{79} (\bibinfo{date}{July} \bibinfo{year}{2019}).
\newblock


\bibitem[\protect\citeauthoryear{Sun, Xu, Zhang, Fanello, Rhemann, Debevec,
  Tsai, Barron, and Ramamoorthi}{Sun et~al\mbox{.}}{2020}]%
        {sun2020light}
\bibfield{author}{\bibinfo{person}{Tiancheng Sun}, \bibinfo{person}{Zexiang
  Xu}, \bibinfo{person}{Xiuming Zhang}, \bibinfo{person}{Sean Fanello},
  \bibinfo{person}{Christoph Rhemann}, \bibinfo{person}{Paul Debevec},
  \bibinfo{person}{Yun-Ta Tsai}, \bibinfo{person}{Jonathan~T. Barron}, {and}
  \bibinfo{person}{Ravi Ramamoorthi}.} \bibinfo{year}{2020}\natexlab{}.
\newblock \showarticletitle{Light Stage Super-Resolution: Continuous
  High-Frequency Relighting}. In \bibinfo{booktitle}{\emph{ACM Trans. on Graph.
  (Proceedings of SIGGRAPH Asia)}}.
\newblock


\bibitem[\protect\citeauthoryear{Tewari, Elgharib, Bharaj, Bernard, Seidel,
  P{\'e}rez, Z{\"o}llhofer, and Theobalt}{Tewari et~al\mbox{.}}{2020a}]%
        {tewari2020stylerig}
\bibfield{author}{\bibinfo{person}{Ayush Tewari}, \bibinfo{person}{Mohamed
  Elgharib}, \bibinfo{person}{Gaurav Bharaj}, \bibinfo{person}{Florian
  Bernard}, \bibinfo{person}{Hans-Peter Seidel}, \bibinfo{person}{Patrick
  P{\'e}rez}, \bibinfo{person}{Michael Z{\"o}llhofer}, {and}
  \bibinfo{person}{Christian Theobalt}.} \bibinfo{year}{2020}\natexlab{a}.
\newblock \showarticletitle{StyleRig: Rigging StyleGAN for 3D Control over
  Portrait Images, CVPR 2020}. In \bibinfo{booktitle}{\emph{Computer Vision and
  Pattern Recognition (CVPR)}}.
\newblock


\bibitem[\protect\citeauthoryear{Tewari, Elgharib, BR, Bernard, Seidel,
  P{\'e}rez, Z{\"o}llhofer, and Theobalt}{Tewari et~al\mbox{.}}{2020b}]%
        {tewari2020pie}
\bibfield{author}{\bibinfo{person}{Ayush Tewari}, \bibinfo{person}{Mohamed
  Elgharib}, \bibinfo{person}{Mallikarjun BR}, \bibinfo{person}{Florian
  Bernard}, \bibinfo{person}{Hans-Peter Seidel}, \bibinfo{person}{Patrick
  P{\'e}rez}, \bibinfo{person}{Michael Z{\"o}llhofer}, {and}
  \bibinfo{person}{Christian Theobalt}.} \bibinfo{year}{2020}\natexlab{b}.
\newblock \showarticletitle{PIE: Portrait Image Embedding for Semantic
  Control}.
\newblock \bibinfo{journal}{\emph{ACM Trans. on Graph. (Proceedings SIGGRAPH
  Asia)}} \bibinfo{volume}{39}, \bibinfo{number}{6}.
\newblock


\bibitem[\protect\citeauthoryear{Tewari, Fried, Thies, Sitzmann, Lombardi,
  Sunkavalli, Martin-Brualla, Simon, Saragih, Nie{\ss}ner,
  et~al\mbox{.}}{Tewari et~al\mbox{.}}{2020c}]%
        {tewari2020state}
\bibfield{author}{\bibinfo{person}{Ayush Tewari}, \bibinfo{person}{Ohad Fried},
  \bibinfo{person}{Justus Thies}, \bibinfo{person}{Vincent Sitzmann},
  \bibinfo{person}{Stephen Lombardi}, \bibinfo{person}{Kalyan Sunkavalli},
  \bibinfo{person}{Ricardo Martin-Brualla}, \bibinfo{person}{Tomas Simon},
  \bibinfo{person}{Jason Saragih}, \bibinfo{person}{Matthias Nie{\ss}ner},
  {et~al\mbox{.}}} \bibinfo{year}{2020}\natexlab{c}.
\newblock \showarticletitle{State of the art on neural rendering}. In
  \bibinfo{booktitle}{\emph{Computer Graphics Forum}},
  Vol.~\bibinfo{volume}{39}. Wiley Online Library, \bibinfo{pages}{701--727}.
\newblock


\bibitem[\protect\citeauthoryear{Thies, Zollh{\"o}fer, and Nie{\ss}ner}{Thies
  et~al\mbox{.}}{2019}]%
        {thies2019deferred}
\bibfield{author}{\bibinfo{person}{Justus Thies}, \bibinfo{person}{Michael
  Zollh{\"o}fer}, {and} \bibinfo{person}{Matthias Nie{\ss}ner}.}
  \bibinfo{year}{2019}\natexlab{}.
\newblock \showarticletitle{Deferred neural rendering: Image synthesis using
  neural textures}.
\newblock \bibinfo{journal}{\emph{ACM Transactions on Graphics (TOG)}}
  \bibinfo{volume}{38}, \bibinfo{number}{4} (\bibinfo{year}{2019}),
  \bibinfo{pages}{1--12}.
\newblock


\bibitem[\protect\citeauthoryear{Wang, Yu, Lu, Wang, Qian, and Xu}{Wang
  et~al\mbox{.}}{2020}]%
        {sipr_ex}
\bibfield{author}{\bibinfo{person}{Zhibo Wang}, \bibinfo{person}{Xin Yu},
  \bibinfo{person}{Ming Lu}, \bibinfo{person}{Quan Wang}, \bibinfo{person}{Chen
  Qian}, {and} \bibinfo{person}{Feng Xu}.} \bibinfo{year}{2020}\natexlab{}.
\newblock \showarticletitle{Single Image Portrait Relighting via Explicit
  Multiple Reflectance Channel Modeling}.
\newblock \bibinfo{journal}{\emph{ACM Trans. on Graph. (Proceedings of SIGGRAPH
  Asia)}} \bibinfo{volume}{39}, \bibinfo{number}{6}, Article
  \bibinfo{articleno}{220} (\bibinfo{year}{2020}).
\newblock


\bibitem[\protect\citeauthoryear{Weyrich, Matusik, Pfister, Bickel, Donner, Tu,
  McAndless, Lee, Ngan, Jensen, and Gross}{Weyrich et~al\mbox{.}}{2006}]%
        {Weyrich2006Analysis}
\bibfield{author}{\bibinfo{person}{Tim Weyrich}, \bibinfo{person}{Wojciech
  Matusik}, \bibinfo{person}{Hanspeter Pfister}, \bibinfo{person}{Bernd
  Bickel}, \bibinfo{person}{Craig Donner}, \bibinfo{person}{Chien Tu},
  \bibinfo{person}{Janet McAndless}, \bibinfo{person}{Jinho Lee},
  \bibinfo{person}{Addy Ngan}, \bibinfo{person}{Henrik~Wann Jensen}, {and}
  \bibinfo{person}{Markus Gross}.} \bibinfo{year}{2006}\natexlab{}.
\newblock \showarticletitle{Analysis of Human Faces using a Measurement-Based
  Skin Reflectance Model}.
\newblock \bibinfo{journal}{\emph{ACM Trans. on Graphics (Proceedings of
  SIGGRAPH)}} \bibinfo{volume}{25}, \bibinfo{number}{3} (\bibinfo{year}{2006}),
  \bibinfo{pages}{1013–1024}.
\newblock


\bibitem[\protect\citeauthoryear{Wiles, Koepke, and Zisserman}{Wiles
  et~al\mbox{.}}{2018}]%
        {Wiles18X2Face}
\bibfield{author}{\bibinfo{person}{Olivia Wiles}, \bibinfo{person}{A.~Sophia
  Koepke}, {and} \bibinfo{person}{Andrew Zisserman}.}
  \bibinfo{year}{2018}\natexlab{}.
\newblock \showarticletitle{X2Face: A network for controlling face generation
  using images, audio, and pose codes}. In \bibinfo{booktitle}{\emph{European
  Conference on Computer Vision (ECCV)}}.
\newblock


\bibitem[\protect\citeauthoryear{Yamaguchi, Saito, Nagano, Zhao, Chen,
  Olszewski, Morishima, and Li}{Yamaguchi et~al\mbox{.}}{2018}]%
        {yamaguchi_high-fidelity_2018}
\bibfield{author}{\bibinfo{person}{Shuco Yamaguchi}, \bibinfo{person}{Shunsuke
  Saito}, \bibinfo{person}{Koki Nagano}, \bibinfo{person}{Yajie Zhao},
  \bibinfo{person}{Weikai Chen}, \bibinfo{person}{Kyle Olszewski},
  \bibinfo{person}{Shigeo Morishima}, {and} \bibinfo{person}{Hao Li}.}
  \bibinfo{year}{2018}\natexlab{}.
\newblock \showarticletitle{High-fidelity facial reflectance and geometry
  inference from an unconstrained image}.
\newblock \bibinfo{journal}{\emph{ACM Trans. on Graph. (Proceedings of
  SIGGRAPH)}} \bibinfo{volume}{37}, \bibinfo{number}{4}, Article
  \bibinfo{articleno}{162} (\bibinfo{year}{2018}).
\newblock


\bibitem[\protect\citeauthoryear{Yang, Chen, Lin, and Chuang}{Yang
  et~al\mbox{.}}{2019}]%
        {yang2019fsa}
\bibfield{author}{\bibinfo{person}{Tsun-Yi Yang}, \bibinfo{person}{Yi-Ting
  Chen}, \bibinfo{person}{Yen-Yu Lin}, {and} \bibinfo{person}{Yung-Yu Chuang}.}
  \bibinfo{year}{2019}\natexlab{}.
\newblock \showarticletitle{Fsa-net: Learning fine-grained structure
  aggregation for head pose estimation from a single image}. In
  \bibinfo{booktitle}{\emph{Proceedings of the IEEE/CVF Conference on Computer
  Vision and Pattern Recognition}}. \bibinfo{pages}{1087--1096}.
\newblock


\bibitem[\protect\citeauthoryear{Zhang, Isola, Efros, Shechtman, and
  Wang}{Zhang et~al\mbox{.}}{2018}]%
        {zhang2018perceptual}
\bibfield{author}{\bibinfo{person}{Richard Zhang}, \bibinfo{person}{Phillip
  Isola}, \bibinfo{person}{Alexei~A Efros}, \bibinfo{person}{Eli Shechtman},
  {and} \bibinfo{person}{Oliver Wang}.} \bibinfo{year}{2018}\natexlab{}.
\newblock \showarticletitle{The Unreasonable Effectiveness of Deep Features as
  a Perceptual Metric}. In \bibinfo{booktitle}{\emph{CVPR}}.
\newblock


\bibitem[\protect\citeauthoryear{Zhang, Barron, Tsai, Pandey, Zhang, Ng, and
  Jacobs}{Zhang et~al\mbox{.}}{2020}]%
        {zhang2020portrait}
\bibfield{author}{\bibinfo{person}{Xuaner Zhang}, \bibinfo{person}{Jonathan~T.
  Barron}, \bibinfo{person}{Yun-Ta Tsai}, \bibinfo{person}{Rohit Pandey},
  \bibinfo{person}{Xiuming Zhang}, \bibinfo{person}{Ren Ng}, {and}
  \bibinfo{person}{David~E. Jacobs}.} \bibinfo{year}{2020}\natexlab{}.
\newblock \showarticletitle{Portrait Shadow Manipulation}.
\newblock \bibinfo{journal}{\emph{ACM Transactions on Graphics (TOG)}}
  \bibinfo{volume}{39}, \bibinfo{number}{4}.
\newblock


\bibitem[\protect\citeauthoryear{Zhou, Hadap, Sunkavalli, and Jacobs}{Zhou
  et~al\mbox{.}}{2019}]%
        {Zhou_2019_ICCV}
\bibfield{author}{\bibinfo{person}{Hao Zhou}, \bibinfo{person}{Sunil Hadap},
  \bibinfo{person}{Kalyan Sunkavalli}, {and} \bibinfo{person}{David~W.
  Jacobs}.} \bibinfo{year}{2019}\natexlab{}.
\newblock \showarticletitle{Deep Single-Image Portrait Relighting}. In
  \bibinfo{booktitle}{\emph{International Conference on Computer Vision
  (ICCV)}}.
\newblock


\bibitem[\protect\citeauthoryear{{Zhou Wang}, {Bovik}, {Sheikh}, and
  {Simoncelli}}{{Zhou Wang} et~al\mbox{.}}{2004}]%
        {ssim2004}
\bibfield{author}{\bibinfo{person}{{Zhou Wang}}, \bibinfo{person}{Alan~C.
  {Bovik}}, \bibinfo{person}{Hamid~R. {Sheikh}}, {and} \bibinfo{person}{Eero~P.
  {Simoncelli}}.} \bibinfo{year}{2004}\natexlab{}.
\newblock \showarticletitle{Image quality assessment: from error visibility to
  structural similarity}.
\newblock \bibinfo{journal}{\emph{IEEE Transactions on Image Processing}}
  \bibinfo{volume}{13}, \bibinfo{number}{4} (\bibinfo{year}{2004}),
  \bibinfo{pages}{600--612}.
\newblock


\end{thebibliography}
\end{document}